\documentclass[conference]{IEEEtran}
\IEEEoverridecommandlockouts
\usepackage{cite}
\usepackage{amsmath,amssymb,amsfonts}
\usepackage{algorithmic}
\usepackage{graphicx}
\usepackage{textcomp}
\usepackage{xcolor}
\def\BibTeX{{\rm B\kern-.05em{\sc i\kern-.025em b}\kern-.08em
    T\kern-.1667em\lower.7ex\hbox{E}\kern-.125emX}}

\usepackage{listings}
\usepackage{subcaption}
\usepackage{caption}
\usepackage{url}
\usepackage{enumitem} 
\newtheorem{remark}{Remark}
\newtheorem{definition}{Definition}
\usepackage{soul}

\begin{document}
\def\pdataset{\mathcal{V}}
\def\dataset{\pdataset}
\def\singleattribute{a}
\def\singledomain{\mathcal{V}}
\def\nsamples{n}
\def\exampleindex{i}
\def\anexample{x}
\def\nfeatures{d}
\def\attributeindex{k}
\def\datasetsindex{g}
\def\ruleindex{j}
\def\ruleindexter{h}
\def\ruleindexbis{l}
\def\branchindex{j}
\def\origd{Orig} 

\def\onecell[#1,#2,#3]{#3_{#1,#2}} 
\def\singlevalue[#1,#2,#3]{v^{#3}_{#1,#2}}
\def\singlevaluetwoargs[#1,#2]{v_{#1,#2}}

\def\generalizedpdataset{\mathcal{W}}
\def\world{w}
\def\worldvar[#1,#2,#3]{\world_{#1,#2}}
\def\possibleworlds{\Pi}

\def\interpretablemodel{M} 
\def\rulelist{RL} 
\def\decisiontree{DT} 
\def\antecedent{f}
\def\consequent{v}
\def\arule[#1]{(\antecedent_{#1}, \consequent_{#1})}
\def\rulelistlength{r'}
\def\nbbranches{r}
\def\branch{f}

\def\capt[#1]{\mathsf{Capt}_{#1}}
\def\numb{\mathsf{num}}
\def\nbexclassified{C}

\def\shannonentropy{H}
\def\oldmetric{\mathsf{Dist}}
\def\newmetric{\mathsf{Dist}_G}

\def\corels{\texttt{CORELS}}
\def\cart{\texttt{CART}}
\def\greedyRL{\texttt{GreedyRL}}
\def\exactdt{\texttt{DL8.5}}
\def\heuristicdt{\texttt{sklearn\_DT}}
\def\sklearn{\texttt{scikit-learn}}
\def\DecisionTreeClassifier{\texttt{DecisionTreeClassifier}}

\lstdefinelanguage{RuleListsLanguage}{
  keywords={if, then, else},
  keywordstyle=\color{blue}\bfseries,
  ndkeywords={},
  ndkeywordstyle=\color{darkgray}\bfseries,
  identifierstyle=\color{black},
  sensitive=false,
  comment=[l]{//},
  morecomment=[s]{/*}{*/},
  commentstyle=\color{purple}\ttfamily,
  stringstyle=\color{red}\ttfamily,
  morestring=[b]',
  morestring=[b]"
}

\definecolor{RuleListsLanguageBackgroundColor}{rgb}{0.95, 0.95, 0.95}

\newcommand{\multilinecomment}[1]{}
\def\mjo#1{\textcolor{cyan}{#1}}
\def\mo#1{\textcolor{red}{#1}}
\def\ulr#1{\textcolor{violet}{#1}}
\def\seb#1{\textcolor{brown}{#1}}
\def\jul#1{\textcolor{orange}{#1}}
\def\revision#1{\textcolor{black}{#1}}
\def\revisionmultiline{\color{black}}

\title{Probabilistic Dataset Reconstruction from Interpretable Models\vspace{10pt}}

\author{\IEEEauthorblockN{1\textsuperscript{st} Julien Ferry}
\IEEEauthorblockA{
\textit{LAAS-CNRS, Universit\'{e} de Toulouse, CNRS}\\
Toulouse, France \\
jferry@laas.fr}
\and
\IEEEauthorblockN{2\textsuperscript{nd} Ulrich A{\"i}vodji}
\IEEEauthorblockA{\textit{\'Ecole de Technologie Sup\'erieure} \\
Montr\'eal, Canada \\
Ulrich.Aivodji@etsmtl.ca}
\and
\IEEEauthorblockN{3\textsuperscript{rd} S\'ebastien Gambs}
\IEEEauthorblockA{\textit{Universit\'e du Qu\'ebec \`a Montr\'eal} \\
Montr\'eal, Canada \\
gambs.sebastien@uqam.ca}
\and
\IEEEauthorblockN{4\textsuperscript{th} Marie-Jos\'e Huguet}
\IEEEauthorblockA{
\textit{LAAS-CNRS, Universit\'{e} de Toulouse, CNRS, INSA}\\
Toulouse, France \\
huguet@laas.fr}
\and
\IEEEauthorblockN{5\textsuperscript{th} Mohamed Siala}
\IEEEauthorblockA{
\textit{LAAS-CNRS, Universit\'{e} de Toulouse, CNRS, INSA}\\
Toulouse, France \\
msiala@laas.fr}
}

\maketitle

\begin{abstract}
Interpretability is often pointed out as a key requirement for trustworthy machine learning. 
However, learning and releasing models that are \emph{inherently interpretable} leaks information regarding the underlying training data. 
As such disclosure may directly conflict with privacy, a precise quantification of the privacy impact of such breach is a fundamental problem.
For instance, previous work~\cite{DBLP:conf/dbsec/GambsGH12} have shown that the structure of a decision tree can be leveraged to build a probabilistic reconstruction of its training dataset, with the uncertainty of the reconstruction being a relevant metric for the information leak.
In this paper, we propose of a novel framework generalizing these probabilistic reconstructions in the sense that it can handle other forms of interpretable models and more generic types of knowledge. 
In addition, we demonstrate that under realistic assumptions regarding the interpretable models' structure, 
the uncertainty of the reconstruction can be computed efficiently.
Finally, we illustrate the applicability of our approach on both decision trees and rule lists, by comparing the theoretical information leak associated to either exact or heuristic learning algorithms. 
Our results suggest that optimal interpretable models are often more compact and leak less information regarding their training data than greedily-built ones, for a given accuracy level.
\end{abstract}


\section{Introduction}

The growing deployment of machine learning systems in real-world decision making systems raises several ethical concerns.
Among them, transparency is often pointed out as a fundamental requirement.
On the one hand, some approaches consist in explaining \emph{black-box} models, whose internals are either hidden or too complex to be understood by the user. 
Such \emph{post-hoc explainability} techniques~\cite{guidotti2018survey} include global explanations~\cite{craven1995extracting,lakkaraju2017interpretable} approximating the entire black-box and local explanations~\cite{ribeiro2016should,lundberg2017unified} of individual predictions. 
One important drawback with these approaches is that they may be misleading by not reflecting the actual black-box model behavior, and can even be manipulated by malicious entities~\cite{aivodji2019fairwashing,slack2020fooling,dimanov2020you,laberge2022fooling,aivodji2021characterizing}.
On the other hand, it is possible to learn inherently \emph{interpretable models}, whose structure can be understood by human users. 
Common types of interpretable models~\cite{freitas2014comprehensible} include rule lists~\cite{DBLP:journals/ml/Rivest87,angelino2017learning}, rule sets~\cite{rijnbeek2010finding} and decision trees~\cite{DBLP:books/wa/BreimanFOS84}.

Another key requirement for the deployment of machine learning models is privacy. 
Indeed, such models are often trained on large amounts of personal data, and it is necessary to ensure that they learn useful generic patterns without leaking private information about individuals. 
In this context, \emph{inference attacks}~\cite{doi:10.1146/annurev-statistics-060116-054123,DBLP:journals/corr/abs-2007-07646,DBLP:journals/corr/abs-2005-08679} leverage the output of a computation (\emph{e.g.}, a trained model) to retrieve information regarding its inputs (\emph{e.g.}, a training dataset). 
Among other types of inferences attacks, \emph{reconstruction attacks}~\cite{doi:10.1146/annurev-statistics-060116-054123,DBLP:journals/corr/abs-2007-07646,DBLP:journals/corr/abs-2005-08679}, aim at reconstructing partly or entirely a model's training set.

While releasing interpretable models can be desirable from a transparency perspective, it intrinsically leaks information regarding the model's training data.
For instance, previous work~\cite{DBLP:conf/dbsec/GambsGH12} exploited this information to build a \emph{probabilistic} reconstruction of a decision tree's training set - effectively implementing a form of \emph{reconstruction attack}.
It is then possible to quantify the amount of information leaked by the model by measuring the uncertainty remaining within the reconstructed probabilistic dataset.
Furthermore, while reconstructions are often not unique, the proposed representation is able to encode a whole set of possible ones.
However, the approach relies on strong assumptions, such as statistical independence and uniform distribution of the random variables modeling the probabilistic dataset. 
While it allows probabilistic reconstructions from decision trees, it is not generic enough to encode more general types of knowledge, and cannot be used with other types of interpretable models, such as rule lists.
In this work, we generalize the notion of probabilistic dataset by relaxing the aforementioned assumptions. 
In particular, we show how the success of such generalized probabilistic reconstructions can be assessed, and illustrate it theoretically and empirically on several forms of interpretable models.
Our contributions can be summarized as follows:
\begin{itemize}[leftmargin=*]
    \item We generalize 
    probabilistic datasets to represent any type of knowledge acquired from an interpretable model's structure.
    \item We extend the metric for quantifying the success of probabilistic reconstruction attacks in a more generic setting.
    \item Although computationally expensive in the generic setting, we show how the success of a probabilistic reconstruction attack may be decomposed under realistic assumptions regarding the structure of the interpretable model considered.
    \item We demonstrate that in the specific case of decision trees and rule lists, the success of a probabilistic reconstruction attack can be estimated efficiently, and theoretically compare the reconstruction quality between these two hypothesis classes.
    \item We implement the proposed approach and compare the reconstruction quality from optimal and heuristically-built models, for both decision trees and rule lists.
\end{itemize}
The outline of the paper is as follows. 
First in Section~\ref{sec_relatedwork}, we review related works on reconstruction attacks before introducing the notion of probabilistic dataset reconstruction. 
In Section~\ref{sec:generalization_pdatasets_metrics}, we generalize this notion to handle more generic forms of knowledge. 
We show in Section~\ref{sec:reconstruction_in_practice} how the success of such generalized probabilistic reconstructions from interpretable models can be assessed.
Finally, we illustrate the applicability of the approach by demonstrating how it can be used to compare the amount of information optimal models carry compared to greedily-built ones.

\section{Related Work}
\label{sec_relatedwork}

In this section, we first provide a brief overview of the existing work with respect to reconstruction attacks.
As this term has been used in the literature to encompass a wide variety of problems and approaches, we then focus on the state-of-the-art on probabilistic datasets reconstruction from interpretable models.

\subsection{Reconstruction Attacks}

Ensuring that the output of a computation over a dataset $\pdataset$ cannot be used to retrieve private information about it is a fundamental objective in privacy~\cite{DBLP:conf/pods/DinurN03}. 
In particular, inference attacks~\cite{doi:10.1146/annurev-statistics-060116-054123,DBLP:journals/corr/abs-2007-07646} aim at retrieving information regarding $\pdataset$ by only observing the outputs of the computation.
In machine learning, such computation is usually a learning algorithm whose output is a trained model.
Two distinct adversarial settings are generally considered~\cite{DBLP:journals/corr/abs-2005-08679,DBLP:journals/corr/abs-2007-07646}. 
In the \emph{black-box setting}, the adversary does not know the model's internals and can only query it through an API. 
In the \emph{white-box setting}, the adversary has full knowledge of the model parameters. 
Diverse types of inference attacks have been proposed against machine learning models~\cite{DBLP:journals/corr/abs-2005-08679}. 
In this paper, we focus on \emph{reconstruction attacks}~\cite{doi:10.1146/annurev-statistics-060116-054123,DBLP:journals/corr/abs-2007-07646,DBLP:journals/corr/abs-2005-08679}, 
in which an adversary aims at recovering parts of a model's training data.


Reconstruction attacks have first been studied in the context of database access mechanisms.
In this setup, a database contains records about individuals, with each record being composed of non-private information along with a private bit~\cite{doi:10.1146/annurev-statistics-060116-054123}. 
The adversary performs queries to a database access mechanism, whose outputs are aggregate and noisy statistics about private bits of individuals in the database.
An efficient linear program for reconstructing private bits of the database leveraging counting queries was proposed~\cite{DBLP:conf/pods/DinurN03} and later improved and generalized to handle other types of queries~\cite{10.1145/1250790.1250804}.
The practical effectiveness of such approaches was demonstrated in several real-world applications~\cite{DBLP:journals/jpc/CohenN20,DBLP:conf/uss/GadottiHRLM19,10.1145/3291276.3295691}.

Other previous works have also tackled reconstruction problems in other settings. 
For example in the pharmacogenetics field, machine learning models predict medical treatments specific to a patient's genotype and background. 
In this sensitive context, a reconstruction attack was proposed, taking as input a trained model and some demographic (non-private) information about a patient whose records were used for training and predicting its sensitive attributes~\cite{DBLP:conf/uss/FredriksonLJLPR14}.
Subsequent works have designed reconstruction attacks leveraging confidence values output by machine learning models to infer private information about training examples given some information about them~\cite{DBLP:conf/ccs/FredriksonJR15}.
Other works have studied the intended~\cite{DBLP:conf/ccs/SongRS17} and unintended~\cite{DBLP:conf/uss/Carlini0EKS19} training data memorization of machine learning models, along with different ways to exploit it in a white-box or black-box setting.
In collaborative deep learning, it was also shown that an adversarial server can exploit the collected gradient updates to recover parts of the participants' data~\cite{DBLP:conf/atis/PhongA0WM17}.
In addition, in the context of online learning, a reconstruction attack was developed to infer the \emph{updating set} (\emph{i.e.}, newly-collected data used to re-train the deployed model) information using a generative adversarial network leveraging the difference between the model before and after its update~\cite{DBLP:conf/uss/0001B0F020}. 

Finally, recent works have considered the special case of training set \emph{sensitive attributes} reconstruction in the context of fair learning~\cite{hu2020inference,aalmoes2022dikaios,hamman2022can,ferry2023exploiting}.
The key challenge here is that while \emph{sensitive attributes} are usually known at training time to ensure the resulting model's fairness, they cannot be used explicitly for inference to avoid disparate treatment~\cite{10.2307/24758720}. 
More precisely, \cite{aalmoes2022dikaios} proposed a machine learning based attack leveraging an auxiliary dataset whose sensitive attributes are known, while~\cite{hu2020inference,ferry2023exploiting} explicitly exploit fairness by encoding it within declarative programming frameworks to enhance the reconstruction. 
Both~\cite{hu2020inference} and~\cite{hamman2022can} consider a particular setup in which a \emph{learner} can query an \emph{auditor} (owning the training set sensitive attributes) to known whether some model's parameters satisfy the fairness constraints. 
They show that the auditor's answers can be leveraged to conduct the attack. 

To the best of our knowledge, the only attack leveraging the structure of a trained interpretable model to build a probabilistic reconstruction of its training set was proposed in~\cite{DBLP:conf/dbsec/GambsGH12}. 
Our \emph{white-box} approach builds upon this baseline work whose key notions are introduced in details hereafter.

\subsection{Probabilistic Dataset Reconstruction from Interpretable Models}
\label{sec:probabilistic_reconstruction_gambs}
\begin{table*}[tb!]
\parbox{.50\linewidth}{
\centering
\captionsetup{justification=centering}
\caption{Example of deterministic dataset $\pdataset^{\origd}$.} 
\label{tab:toy_dataset_orig}
\begin{tabular}{c|c|c|c|c|}
\cline{2-5}
                                     & $\singleattribute_1$ & $\singleattribute_2$ & $\singleattribute_3$ & Label \\ \hline
\multicolumn{1}{|c|}{$\anexample_1$} & 12                   & 0                    & 3                    & 0     \\ \hline
\multicolumn{1}{|c|}{$\anexample_2$} & 14                   & 1                    & 2                    & 0     \\ \hline
\multicolumn{1}{|c|}{$\anexample_3$} & 11                   & 1                    & 2                    & 1     \\ \hline
\multicolumn{1}{|c|}{$\anexample_4$} & 14                   & 0                    & 1                    & 1     \\ \hline
\end{tabular}
}
\hfill
\parbox{.50\linewidth}{
\centering
\captionsetup{justification=centering}
\caption{Example of probabilistic dataset $\pdataset^{\decisiontree}$ reconstructed from a Decision Tree (Figure~\ref{fig:example_toy_dt}).} \label{tab:toy_dataset_rec}
\begin{tabular}{c|c|c|c|c|}
\cline{2-5}
                                     & $\singleattribute_1$              & $\singleattribute_2$ & $\singleattribute_3$ & Label \\ \hline
\multicolumn{1}{|c|}{$\anexample_1$} & $ \in \{12, 13, 14, 15\}$         & $ \in \{0, 1\}$      & $ \in \{2, 3\}$      & 0     \\ \hline
\multicolumn{1}{|c|}{$\anexample_2$} & $ \in \{12, 13, 14, 15\}$         & $ \in \{0, 1\}$      & $ \in \{2, 3\}$      & 0     \\ \hline
\multicolumn{1}{|c|}{$\anexample_3$} & $ \in \{10, 11\}$                 & $ \in \{0, 1\}$      & $ \in \{2, 3\}$      & 1     \\ \hline
\multicolumn{1}{|c|}{$\anexample_4$} & $ \in \{10, 11, 12, 13, 14, 15\}$ & $ \in \{0, 1\}$      & $ \in \{1\}$         & 1     \\ \hline
\end{tabular}
}
\hfill
\hfill
\end{table*}

In~\cite{DBLP:conf/dbsec/GambsGH12}, the structure of a decision tree is used to build a probabilistic reconstruction of its training dataset, in the form of a \emph{probabilistic dataset}, as introduced in Definition~\ref{def:probabilistic_datasets}.

\begin{definition} \label{def:probabilistic_datasets} \textbf{(Probabilistic Dataset)~\cite{DBLP:conf/dbsec/GambsGH12}.} 
A probabilistic dataset $\pdataset$ is composed of $\nsamples$ \emph{examples} $\{\anexample_1,\ldots,\anexample_\nsamples\}$ (the dataset's rows), each consisting in a vector of $\nfeatures$ attributes $\{\singleattribute_1,\ldots,\singleattribute_\nfeatures\}$ (the dataset's columns).
Each attribute $\singleattribute_\attributeindex$ has a domain of definition $\singledomain_\attributeindex$ that includes all the possible values of this attribute. 
The knowledge about attribute $\singleattribute_\attributeindex$ of example $\anexample_\exampleindex$ is modeled by a probability distribution over all the possible values of this attribute, using random variable $\onecell[\exampleindex,\attributeindex,\pdataset{}]$.
Importantly, variables $\{\onecell[\exampleindex \in [1..\nsamples],\attributeindex \in [1\ldots\nfeatures],\pdataset]\}$ are assumed to be statistically independent from each other and their probability distribution to be uniform.
\end{definition}

In practice, if a particular value $\singlevaluetwoargs[\exampleindex,\attributeindex] \in \singledomain_\attributeindex$ of an attribute gathers all the probability mass (\emph{i.e.,} it is perfectly determined: $\mathbb{P}(\onecell[\exampleindex,\attributeindex,\pdataset] = \singlevaluetwoargs[\exampleindex,\attributeindex]) = 1$), then the attribute is said to be deterministic. By extension, a probabilistic dataset whose attributes are all deterministic (\emph{i.e.}, the knowledge about the dataset is perfect) is called a \emph{deterministic dataset}.
Previous work~\cite{DBLP:conf/dbsec/GambsGH12} proposes a procedure to build a probabilistic dataset $\pdataset^{\decisiontree}$ given the structure of a trained decision tree $\decisiontree$. 
Such probabilistic dataset gathers the knowledge that the decision tree inherently encodes about its (deterministic) training dataset $\pdataset^{\origd}$. 
The construction of this probabilistic dataset can then be coined as a \emph{probabilistic reconstruction attack}.
By construction, $\pdataset^{\decisiontree}$ is \emph{compatible} with $\pdataset^{\origd}$: the true value $\singlevalue[\exampleindex,\attributeindex,\origd]$ of any attribute $\singleattribute_\attributeindex$ for any example $\anexample_\exampleindex$ is always contained within the set of possible values for this attribute and this example in the probabilistic reconstruction:
$\mathbb{P}\left(\onecell[\exampleindex,\attributeindex,\pdataset^{\decisiontree}] = \singlevalue[\exampleindex,\attributeindex,\origd]\right) > 0$.  
A natural way to quantify the \emph{success of the probabilistic reconstruction attack} is in terms of the average amount of uncertainty that remains in the built probabilistic dataset $\pdataset^{\decisiontree}$, as stated in the following definition.


\begin{definition} 
\label{def:distance_metric} 
\textbf{(Measure of success of a probabilistic reconstruction attack)~\cite{DBLP:conf/dbsec/GambsGH12}.} 
Let $\pdataset^{\origd}$ be a deterministic dataset composed of $\nsamples$ examples and $\nfeatures$ attributes, used to train a machine learning model $\interpretablemodel$.
Let $\pdataset^{\interpretablemodel}$ be a probabilistic dataset reconstructed from $\interpretablemodel$. 
By construction, $\pdataset^{\interpretablemodel}$ is compatible with $\pdataset^{\origd}$. 
The success of the reconstruction is quantified as the average uncertainty reduction over all attributes of all examples in the dataset:
\begin{align}
    \oldmetric(\pdataset^{\interpretablemodel},\pdataset^{\origd}) = 
    \frac{1}{\nsamples \cdot \nfeatures} \sum_{\exampleindex=1}^{\nsamples}{\sum_{\attributeindex=1}^{\nfeatures}{\frac{\shannonentropy(\onecell[\exampleindex,\attributeindex,\pdataset^{\interpretablemodel}])}{\shannonentropy(\onecell[\exampleindex,\attributeindex,\pdataset])}}} \label{equ:dist_definition}
\end{align}
in which random variable $\onecell[\exampleindex,\attributeindex,\pdataset]$ corresponds to an \emph{uninformed} reconstruction, uniformly distributed over all possible values $\singledomain_\attributeindex$ of attribute $\singleattribute_\attributeindex$, and $\shannonentropy$ denotes the Shannon entropy.

\end{definition}

\begin{figure}[h!]
    \centering
    \includegraphics[width=0.35\textwidth]{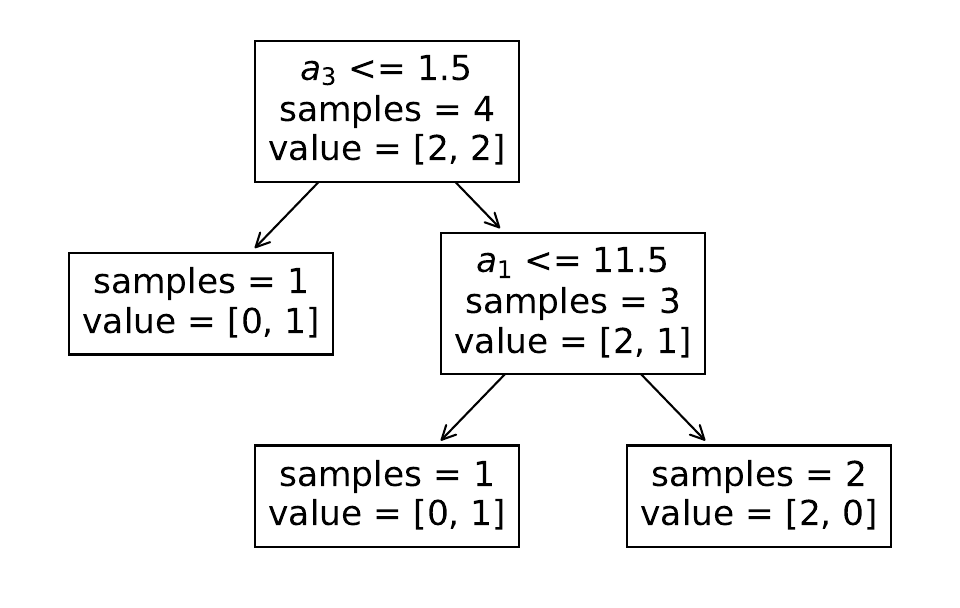}
    \caption{Example of Decision Tree $\decisiontree$ trained using \texttt{scikit-learn}~\cite{scikit-learn}, with $1.0$ accuracy on $\pdataset^{\origd}$ (Table~\ref{tab:toy_dataset_orig}).}
    \label{fig:example_toy_dt}
\end{figure}

Smaller values of $\oldmetric(\pdataset^{\interpretablemodel},\pdataset^{\origd})$ indicate better reconstruction performance (\emph{i.e.}, a more successful attack). 
In particular, 
if $\pdataset^{\interpretablemodel}=\pdataset^{\origd}$, $\oldmetric(\pdataset^{\interpretablemodel},\pdataset^{\origd})=0$: the reconstruction is perfect and there is no uncertainty at all. 
In contrast, in other extreme in which $\pdataset^{\interpretablemodel}$ contains no knowledge at all, $\oldmetric(\pdataset^{\interpretablemodel},\pdataset^{\origd}) = 1$.
\multilinecomment{
\begin{remark}
The definition of $\oldmetric$ provided in~\cite{DBLP:conf/dbsec/GambsGH12} is slightly different, with the objective of using the metric proposed to compute a ``distance between two probabilistic datasets".
In a nutshell, it computes the average uncertainty reduction (compared to the uninformed reconstruction) when the sets of possible values of each cell of the two probabilistic datasets are merged, assuming a uniform probability distribution over these values.
However, when applied to two probabilistic datasets, this metric does not actually measure the distance between both (\emph{e.g.}, for a dataset with no knowledge at all $\pdataset^{Dumb}$, $\oldmetric(\pdataset^{Dumb},\pdataset^{Dumb}) = 1$). 
Rather, it quantifies, for each example and each attribute, the uncertainty that remains about the particular value of an attribute if the two information are pooled together~\cite{DBLP:conf/dbsec/GambsGH12}.
Furthermore, this metric only leverages the alphabets (sets of possible values) of each random variable (assuming uniform probability distributions). 
In practice, 
our proposed formulation meets the requirements of the original $\oldmetric$, but also extend it to be able to integrate potentially complex probability distributions over attributes' values.
\end{remark}
}

\begin{remark}
Definitions~\ref{def:probabilistic_datasets} and~\ref{def:distance_metric} are slightly more general than in~\cite{DBLP:conf/dbsec/GambsGH12}.
Indeed, both use actual random variables while in~\cite{DBLP:conf/dbsec/GambsGH12} each attribute of each example is simply modeled via a set of possible values, which is only suitable under the assumed hypothesis of statistical independence and uniform distribution of the random variables.
Thus, our extended formulation eases the generalization we further provide in Section~\ref{sec:generalization_pdatasets_metrics} while encompassing this particular case.
\end{remark}

\begin{table*}[tb!]
\parbox{.50\linewidth}{
\centering
\captionsetup{justification=centering}
\caption{Example of deterministic dataset $\pdataset^{\origd'}$.} 
\label{tab:toy_dataset_orig_rl}
\begin{tabular}{c|c|c|c|c|}
\cline{2-5}
                                      & $\singleattribute'_1$ & $\singleattribute'_2$ & $\singleattribute'_3$ & Label \\ \hline
\multicolumn{1}{|c|}{$\anexample'_1$} & 1                     & 1                     & 1                     & 1     \\ \hline
\multicolumn{1}{|c|}{$\anexample'_2$} & 1                     & 1                     & 0                     & 1     \\ \hline
\multicolumn{1}{|c|}{$\anexample'_3$} & 0                     & 1                     & 1                     & 0     \\ \hline
\multicolumn{1}{|c|}{$\anexample'_4$} & 1                     & 0                     & 1                     & 0     \\ \hline
\multicolumn{1}{|c|}{$\anexample'_5$} & 1                     & 0                     & 0                     & 1     \\ \hline
\end{tabular}
}
\hfill
\parbox{.50\linewidth}{
\centering
\captionsetup{justification=centering}
\caption{Example of (generalized) probabilistic dataset $\generalizedpdataset^{\rulelist}$ reconstructed from Rule List~\ref{rl:example}.} \label{tab:toy_dataset_rec_rl}
\begin{tabular}{c|cc|c|c|}
\cline{2-5}
                                      & \multicolumn{1}{c|}{$\singleattribute'_1$} & $\singleattribute'_2$ & $\singleattribute'_3$ & Label \\ \hline
\multicolumn{1}{|c|}{$\anexample'_1$} & \multicolumn{1}{c|}{1}                     & 1                     & $\in \{0,1\}$         & 1     \\ \hline
\multicolumn{1}{|c|}{$\anexample'_2$} & \multicolumn{1}{c|}{1}                     & 1                     & $\in \{0,1\}$         & 1     \\ \hline
\multicolumn{1}{|c|}{$\anexample'_3$} & \multicolumn{2}{c|}{$\in \{(0,0), (0,1), (1,0) \}$}                & 1                     & 0     \\ \hline
\multicolumn{1}{|c|}{$\anexample'_4$} & \multicolumn{2}{c|}{$\in \{(0,0), (0,1), (1,0) \}$}                & 1                     & 0     \\ \hline
\multicolumn{1}{|c|}{$\anexample'_5$} & \multicolumn{2}{c|}{$\in \{(0,0), (0,1), (1,0) \}$}                & 0                     & 1     \\ \hline
\end{tabular}
}
\hfill
\hfill
\end{table*}

We illustrate the reconstruction process proposed in~\cite{DBLP:conf/dbsec/GambsGH12} with a toy example. 
A deterministic dataset $\pdataset^{\origd}$, provided in Table~\ref{tab:toy_dataset_orig} is used to train a decision tree classifier $\decisiontree$ depicted in Figure~\ref{fig:example_toy_dt}. 
This dataset includes four examples $\anexample_{\exampleindex \in \{1,2,3,4\}}$ with three attributes $\singleattribute_{\attributeindex \in \{1,2,3\}}$ with domains $\singledomain_1 = \{10, 11, 12, 13, 14, 15\}$, $\singledomain_2 = \{0, 1\}$ and $\singledomain_3 = \{1, 2, 3\}$.
This decision tree, learnt using the \sklearn{} python library~\cite{scikit-learn}, provides the per-label number of training examples in each internal node and each leaf. 
Intuitively, its structure can then be used to reconstruct a probabilistic version of its training dataset $\pdataset^{\decisiontree}$, given in Table~\ref{tab:toy_dataset_rec}.
The algorithm used to build $\pdataset^{\decisiontree}$ simply follows each branch and performs the domains' reductions associated to each split along the branch.

Using Definition~\ref{def:distance_metric},
we can compute the \emph{success of the reconstruction} as the average amount of uncertainty contained within $\pdataset^{\decisiontree}$. 
For instance, $\singledomain_1 = \{10, 11, 12, 13, 14, 15\}$ and $\onecell[3,1,\pdataset^{\decisiontree}]$ takes values in $\{10, 11\}$.
Then, considering all the possible values as equally probable, the uncertainty reduction for attribute $\singleattribute_1$ of example $\anexample_3$ is:$\frac{\shannonentropy(\onecell[3,1,\pdataset^{\decisiontree}])}{\shannonentropy(\onecell[3,1,\pdataset])} = \frac{-log_{2}(\frac{1}{2})}{-log_{2}(\frac{1}{6})} \approx 0.387$.
By averaging such computation over the entire dataset (\emph{i.e.}, over all attributes of all examples), we obtain $\oldmetric(\pdataset^{\decisiontree},\pdataset^{\origd}) \approx 0.736 $.
To facilitate reading, we aligned $\pdataset^{\origd}$ and $\pdataset^{\decisiontree}$. 
In practice, such alignment can be performed using the Hungarian algorithm~\cite{kuhn1955hungarian,munkres1957algorithms} as is done in~\cite{DBLP:conf/dbsec/GambsGH12}. 
In a nutshell, it consists in performing a minimum cost matching between the examples of $\pdataset^{\origd}$ and those of $\pdataset^{\decisiontree}$, where the assignment cost is computed as the sum of the distances between the paired examples.
Intuitively, the objective is to determine \emph{to which example within $\pdataset^{\origd}$ corresponds each reconstructed example in $\pdataset^{\decisiontree}$}.
However, this would not be needed in scenarios in which $\pdataset^{\origd}$ is unknown, as $\pdataset^{\decisiontree}$ is compatible with $\pdataset^{\origd}$ by construction, and $\oldmetric$ (\ref{equ:dist_definition}) does not require further information regarding $\pdataset^{\origd}$.

Hereafter, we generalize the notions introduced in this section to be able to handle more general type of knowledge, corresponding to other types of interpretable models.

\section{Generalizing Probabilistic Datasets Reconstruction}
\label{sec:generalization_pdatasets_metrics}
In this section, we first illustrate the limits of probabilistic datasets and motivate the need to relax some of their underlying assumptions.
Consequently, 
we introduce \emph{generalized probabilistic datasets}, which 
can be used to encode any arbitrary knowledge regarding a dataset.
Finally, we define a generalized metric $\newmetric$ that can be used to quantify uncertainty reduction within such datasets.

\subsection{Motivation}
\label{sec:motivation_generalization}

The concept of probabilistic dataset as described in Definition~\ref{def:probabilistic_datasets} is suitable to encode knowledge regarding a dataset, as long as this knowledge involves each  \emph{cell} (\emph{i.e.}, which corresponds to one attribute for one example) individually.
For instance, this is appropriate for decision trees in which an example is classified \emph{exactly} by one branch. 
Furthermore, each branch corresponds to a \emph{conjunction} (\emph{i.e.}, logical AND) of conditions (\emph{splits}) over features, which all have to be satisfied.
These conditions allow for the reduction of each such feature's domains \emph{individually}.
However, for other representations of interpretable classifiers, such as rule lists or rule sets, this condition will not be valid.
Again, we illustrate this observation using a toy example.

More precisely, Rule List~\ref{rl:example} was trained on (deterministic) dataset $\pdataset^{\origd'}$, shown in Table~\ref{tab:toy_dataset_orig_rl}. 
It gathers five examples $\anexample'_{\exampleindex \in \{1,2,3,4,5\}}$ described by three binary attributes, named $\singleattribute'_{\attributeindex \in \{1,2,3\}}$ (with domains $\singledomain'_{\attributeindex \in \{1,2,3\}} = \{0,1\}$).
For each rule (including the default rule), $\rulelist$ indicates the number of training examples it captures, for each class. For example, the second rule captures two training examples belonging to class $0$ (here, $\anexample'_{3}$ and $\anexample'_{4}$).

\lstset{numberbychapter=false,caption=\lstname,frame=single, stepnumber=1, numbersep=2pt, xleftmargin=0.05\linewidth, xrightmargin=0.05\linewidth}
\renewcommand{\lstlistingname}{Rule list}
\begin{lstlisting}[captionpos=b,escapeinside={(*}{*)}, language=RuleListsLanguage,backgroundcolor = \color{RuleListsLanguageBackgroundColor}, basicstyle=\footnotesize, caption={Example rule list $\rulelist$ trained using \corels{}~\cite{angelino2017learning,angelino2018learning}, with 1.0 accuracy on $\pdataset^{\origd'}$ (Table~\ref{tab:toy_dataset_orig_rl}).}, label=rl:example]
if [(*$a'_1$*)] and [(*$a'_2$*)] then [true] ([0 ; 2] examples)
else if [(*$a'_3$*)] then [false] ([2 ; 0] example)
else [true] ([0 ; 1] example)
\end{lstlisting} 


The algorithm reconstructing a probabilistic version of a rule list $\rulelist$'s training set from $\rulelist$ itself simply follows the path of each example. 
For an example classified by the $\ruleindex$th rule, it reduces the domains of the attributes involved in the  $\ruleindex$th rule accordingly. It also eliminates all attributes' conjunctions contradicting the fact that the example did not match the previous rules within $\rulelist$.

For instance, the following knowledge can be extracted from Rule List~\ref{rl:example}:
\begin{itemize}
    \item The \textbf{first rule} indicates that for $2$ (positive) examples, the two Boolean attributes $\singleattribute'_1$ and $\singleattribute'_2$ are true.
    \item Using the \textbf{second rule}, we know that the Boolean attribute $\singleattribute'_3$ is true for $2$ (negatively-labelled) examples. \emph{Furthermore, we know that $\singleattribute'_1$ and $\singleattribute'_2$ can not be simultaneously true for these examples (or else they would have been captured by the first rule).} 
    \item Finally, the \textbf{default rule} states that for $1$ (positively-labelled) example, $\singleattribute'_3$ is false, and \emph{$\singleattribute'_1$ and $\singleattribute'_2$ can not be simultaneously true}.
\end{itemize} 

Using such knowledge, one can build a (generalized) probabilistic dataset as shown in Table~\ref{tab:toy_dataset_rec_rl}. 
In this example, part of the model's knowledge directly reduces the individual domains of some attributes for the concerned examples.
As such, the information it brings will successfully be quantified by $\oldmetric$ and encoded in a probabilistic dataset.
However, other information (specified \emph{in italic}) does not reduce any attribute's domain individually.
For instance, as shown in Table~\ref{tab:toy_dataset_rec_rl}, one knows that for examples $\anexample'_{3}$ and $\anexample'_{4}$,  $\singleattribute'_1$ and $\singleattribute'_2$ can not simultaneously be true. 
Nevertheless, taken apart, their respective domains would be unchanged as both binary attributes can still take values in $\{0,1\}$.
While such knowledge brings information for reconstruction, this cannot be quantified using $\oldmetric$ nor represented using a probabilistic dataset as formalized in Definition~\ref{def:probabilistic_datasets}.

Indeed, one key assumption with 
Definition~\ref{def:probabilistic_datasets} is that the random variables representing each attribute for each example are independent from each other. 
This is leveraged by $\oldmetric$, which computes the reductions of the individual entropies.
However, this representation cannot handle more generic knowledge, in which uncertainty can be spread jointly across multiple random variables.
This limitation is also pointed out in the theory of probabilistic databases.
More precisely, quoting~\cite{DBLP:series/synthesis/2011Suciu}, this representation (talking about a scheme similar to probabilistic datasets as formalized in Definition~\ref{def:probabilistic_datasets} and illustrated in Figure~\ref{tab:toy_dataset_rec}) is ``more compact'', as we do not need to expand all possible combinations of the different variables' values explicitly. However, ``it cannot account for correlations across possible readings of different fields, such as when we know that no two persons can have the same social security number''. In this particular case, this corresponds to a correlation across examples, while in the aforementioned example of Rule List~\ref{rl:example} we observed correlations between attributes within the same example. For instance, to encode the knowledge regarding attributes $\singleattribute'_1$ and $\singleattribute'_2$ of examples $\anexample'_{3}$ and $\anexample'_{4}$, we had to enumerate all the possible combinations of these two attributes' values (Table~\ref{tab:toy_dataset_rec_rl}).
    
\subsection{Generalized Probabilistic Datasets}

As illustrated in the previous subsection, the assumptions underlying probabilistic datasets (Definition~\ref{def:probabilistic_datasets}) - namely statistical independence and uniform distribution of their random variables - make them inappropriate in the general case.
Generalized probabilistic datasets remove these assumptions as stated in Definition~\ref{def:generalized_probabilistic_datasets}.


\begin{definition} \label{def:generalized_probabilistic_datasets} \textbf{(Generalized probabilistic dataset).} 
A generalized probabilistic dataset $\generalizedpdataset$ 
is composed of $\nsamples$ \emph{examples} $\{\anexample_1,\ldots,\anexample_\nsamples\}$ (the dataset's rows), each consisting in a vector of $\nfeatures$ attributes $\{\singleattribute_1,\ldots,\singleattribute_\nfeatures\}$ (the dataset's columns).
The knowledge about attribute $\singleattribute_\attributeindex$ of example $\anexample_\exampleindex$ is modeled by a probability distribution over all the possible values of this attribute, using random variable $\onecell[\exampleindex,\attributeindex,\generalizedpdataset{}]$.
Importantly, variables $\{\onecell[\exampleindex \in [1..\nsamples],\attributeindex \in [1..\nfeatures],\generalizedpdataset]\}$ are not necessarily statistically independent from each other and can follow any arbitrary distribution.
Each possible instantiation $\world = \{ \worldvar[\exampleindex \in [1..\nsamples],\attributeindex \in [1..\nfeatures],]\}$ of the $\onecell[\exampleindex \in [1..\nsamples],\attributeindex \in [1..\nfeatures],\generalizedpdataset]$ variables (\emph{i.e.}, each deterministic dataset compatible with $\generalizedpdataset$) is named a \emph{possible world}.
We let $\possibleworlds(\generalizedpdataset)$ denote the set of possible worlds within $\generalizedpdataset$: $\possibleworlds(\generalizedpdataset) = \{ \world ~|~ \mathbb{P}(\onecell[\exampleindex \in [1..\nsamples],\attributeindex \in [1..\nfeatures],\generalizedpdataset] = \worldvar[\exampleindex \in [1..\nsamples],\attributeindex \in [1..\nfeatures],]) > 0\}$.
\end{definition}

Again, if all its variables are determined, a generalized probabilistic dataset is said to deterministic.
A key difference between probabilistic datasets and their generalized counterparts is that the set of possible worlds of a probabilistic dataset simply consists in all combinations of the possible variables' values, all random variables being statistically independent.
For generalized probabilistic datasets, it is not the case as there can exist complex inter-dependencies between the random variables that directly influence $\possibleworlds(\generalizedpdataset)$ (as illustrated in Section~\ref{sec:motivation_generalization}). 
Importantly, note that generalized probabilistic datasets as introduced in Definition~\ref{def:generalized_probabilistic_datasets} can encode any knowledge regarding the inferred reconstruction, as they simply consist in a set of random variables $\{\onecell[\exampleindex \in [1..\nsamples],\attributeindex \in [1..\nfeatures],\generalizedpdataset]\}$ with no further assumption regarding their distribution. Furthermore, because these variables can be categorical, but also continuous, there is no particular restriction regarding the type of attributes contained in the reconstructed dataset.

Our generalized probabilistic dataset definition matches the notions of \emph{probabilistic} or \emph{incomplete databases} that are used in the theory of probabilistic databases~\cite{DBLP:series/synthesis/2011Suciu}.
Both correspond to database representations in which some values are uncertain.
More precisely, an \emph{incomplete database} defines a set of \emph{possible worlds}, denoting the possible states of the database (\emph{i.e.}, set of values for the different relations). 
Such \emph{worlds} are also called \emph{relational database instances}, and correspond to all deterministic datasets compatible with our generalized probabilistic dataset in the context of this work. 
If one can associate a probability to each possible world, then the database is called a \emph{probabilistic database} - which
generalizes incomplete databases. 
In the context of this work, one could leverage external knowledge (\emph{e.g.}, demographic information about the data distribution) to associate probabilities to the possible worlds in $\possibleworlds(\generalizedpdataset)$. 
This would lead to a reduction of the uncertainty of the dataset (thus lowering its joint entropy and raising the reconstruction success). 

Both incomplete and probabilistic databases are semantic definitions for which designing a practical representation is challenging~\cite{DBLP:series/synthesis/2011Suciu}. 
Indeed, it is always possible to represent a probabilistic database (here, a generalized probabilistic dataset $\generalizedpdataset$) by explicitly enumerating all its possible worlds (here, $\possibleworlds(\generalizedpdataset)$). 
However, such a mechanism is difficultly applicable when the set of possible worlds is very large.
To circumvent this issue, some \emph{compact} representations have been proposed.
For instance, in \emph{conditional tables} (or \emph{c-tables}),
the different values of the database cells are associated to a propositional formula, called condition, over some random variables. 
The different assignments of the random variables define the different states of the database (\emph{i.e.}, possible worlds). 
\emph{Probabilistic conditional tables} (or \emph{pc-tables}) extend this concept by assigning probabilities to the conditional variables assignments. 
While (p)c-tables may be an interesting representation for generalized probabilistic datasets, we do not assume any specific representation for our generalized probabilistic datasets in this work.
Rather, we demonstrate in Section~\ref{sec:reconstruction_in_practice} that in the context of training set reconstruction from an interpretable model, we can quantify the amount of uncertainty that remains in the resulting generalized probabilistic dataset without building it explicitly (which in practice may be infeasible even with efficient structures such as c-tables).

\subsection{Generalized Measure of the Attack Success}

We now generalize the metric introduced in Definition~\ref{def:distance_metric} to quantify the success of a probabilistic reconstruction attack.
As stated in Definition~\ref{def:distance_metric_generalized}, our new metric $\newmetric$ is more general as it quantifies the uncertainty reduction on the entire dataset using the joint entropy of the underlying random variables. 

\begin{definition} 
\label{def:distance_metric_generalized} \textbf{(Generalized measure of success of a probabilistic reconstruction attack).}
Let $\generalizedpdataset^{\origd}$ be a deterministic dataset composed of $\nsamples$ examples and $\nfeatures$ attributes, used to train a machine learning model $\interpretablemodel$.
Let $\generalizedpdataset^{\interpretablemodel}$ be a generalized probabilistic dataset reconstructed from $\interpretablemodel$. 
By construction, $\generalizedpdataset^{\interpretablemodel}$ is compatible with $\generalizedpdataset^{\origd}$ (\emph{i.e.}, $\generalizedpdataset^{\origd} \in \possibleworlds(\generalizedpdataset^{\interpretablemodel})$). 
The success of the performed reconstruction is quantified as the overall uncertainty reduction in the dataset:
\begin{align}
\newmetric(\generalizedpdataset^{\interpretablemodel},\generalizedpdataset^{\origd}) &= \frac{\shannonentropy\left(\{ \onecell[\exampleindex,\attributeindex,\generalizedpdataset^{\interpretablemodel}] \mid {\exampleindex \in [1..\nsamples], \attributeindex \in [1..\nfeatures]}\}\right)}{\shannonentropy\left(\{ \onecell[\exampleindex,\attributeindex,\generalizedpdataset] \mid \exampleindex \in [1..\nsamples], \attributeindex \in [1..\nfeatures]\}\right)}\label{eq:generalized_metric_line_1}\\
&= \frac{\sum_{\world \in \possibleworlds(\generalizedpdataset^{\interpretablemodel})}{- \mathbb{P}(\world) \cdot log_{2}(\mathbb{P}(\world))}}{\sum_{\exampleindex=1}^{\nsamples}{\sum_{\attributeindex=1}^{\nfeatures}{\shannonentropy(\onecell[\exampleindex,\attributeindex,\generalizedpdataset])}}} \label{eq:generalized_metric_line_2}
\end{align}
in which $\shannonentropy$ denotes the Shannon entropy (or joint entropy, when applied to a set of variables, as in~(\ref{eq:generalized_metric_line_1})), and random variable $\onecell[\exampleindex,\attributeindex,\generalizedpdataset]$ corresponds to an \emph{uninformed} reconstruction, uniformly distributed over all possible values of attribute $\singleattribute_\attributeindex$.
\end{definition}

The denominator in Equation~(\ref{eq:generalized_metric_line_1}) can be decomposed as a sum in Equation~(\ref{eq:generalized_metric_line_2}) because the random variables $\onecell[\exampleindex \in [1..\nsamples], \attributeindex \in [1..\nfeatures],\generalizedpdataset]$ are independent from each other, and the joint entropy of a set of variables is equal to the sum of the individual entropies of the variables in the set if and only if the variables are statistically independent. 
This is not the case for variables $\onecell[\exampleindex \in [1..\nsamples], \attributeindex \in [1..\nfeatures],\generalizedpdataset^{\interpretablemodel}]$, and thus the generalized probabilistic dataset has to be considered as a whole through its set of possible worlds $\possibleworlds(\generalizedpdataset^{\interpretablemodel})$.
Again, note that Definition~\ref{def:distance_metric_generalized} is general enough to quantify the uncertainty reduction brought by any type of knowledge, as no assumption is made regarding the distribution of the generalized probabilistic dataset variables, and $\newmetric$ considers their joint entropy.
Furthermore, to handle continuous attributes, the sum over the set of possible worlds $\possibleworlds(\generalizedpdataset^{\interpretablemodel})$ in Equation~(\ref{eq:generalized_metric_line_2}) can be changed into an integral calculation.

The key properties of $\oldmetric$ also hold for $\newmetric$. 
In particular, for any deterministic dataset $\generalizedpdataset^{\origd}$, we have $\newmetric(\generalizedpdataset^{\origd},\generalizedpdataset^{\origd}) = 0$. 
Furthermore, if $\generalizedpdataset^{\interpretablemodel}$ contains no knowledge at all, we have that $\newmetric(\generalizedpdataset^{\interpretablemodel},\generalizedpdataset^{\origd}) = 1$ for any deterministic dataset $\generalizedpdataset^{\origd}$.

One important difference between $\oldmetric$ and $\newmetric$ is the fact, that due to its averaging over the per-example-per-attribute individual uncertainty reductions, $\oldmetric$ considers all features equal (in terms of contribution to the overall uncertainty) while it is not the case for $\newmetric$.
To illustrate this, let us assume a toy scenario with a (deterministic) dataset $\pdataset^{\origd}$ with a single record $\anexample_1 = (1, 1)$ and two attributes $\singleattribute_1$ and $\singleattribute_2$ with domains $\singledomain_1 = \{0, 1\}$ and $\singledomain_2 = \{1, 2, 3\}$. 
Consider the two probabilistic datasets $\pdataset^{rec1}$, in which we know that for $\anexample_1$, $\singleattribute_1 = 1$, and $\pdataset^{rec2}$, in which we know that for $\anexample_1$, $\singleattribute_2 = 1$. 
These datasets are summarized in Tables~\ref{tab:toy_dataset_normalization_orig},~\ref{tab:toy_dataset_normalization_rec1} and~\ref{tab:toy_dataset_normalization_rec2}.
\begin{table}[h!]
\parbox{.289\linewidth}{
\centering
\captionsetup{justification=centering}
\caption{$\pdataset^{Orig}$} \label{tab:toy_dataset_normalization_orig}
\begin{tabular}{c|c|c|}
\cline{2-3}
                                     & $\singleattribute_1$ & $\singleattribute_2$ \\ \hline
\multicolumn{1}{|c|}{$\anexample_1$} & $1$                  & $1$                  \\ \hline
\end{tabular}
}
\hspace{-7pt}
\parbox{.33\linewidth}{
\centering
\captionsetup{justification=centering}
\caption{$\pdataset^{rec1}$} \label{tab:toy_dataset_normalization_rec1}
\begin{tabular}{c|c|c|}
\cline{2-3}
                                     & $\singleattribute_1$ & $\singleattribute_2$ \\ \hline
\multicolumn{1}{|c|}{$\anexample_1$} & $1$                  & $\in \{1,2,3\}$      \\ \hline
\end{tabular}
}
\hfill
\parbox{.33\linewidth}{
\centering
\captionsetup{justification=centering}
\caption{$\pdataset^{rec2}$} \label{tab:toy_dataset_normalization_rec2}
\begin{tabular}{c|c|c|}
\cline{2-3}
                                     & $\singleattribute_1$ & $\singleattribute_2$ \\ \hline
\multicolumn{1}{|c|}{$\anexample_1$} & $\in \{0, 1\}$       & $1$                  \\ \hline
\end{tabular}
}
\hfill
\end{table}

Using Definition~\ref{def:distance_metric}, we have: $\oldmetric(\pdataset^{rec1},\pdataset^{\origd}) = 0.5$, as: 
{\small  
\begin{align*}\frac{\shannonentropy(\onecell[1,1,\pdataset^{rec1}])}{\shannonentropy(\onecell[1,1,\pdataset])} = \frac{-log_{2}(1)}{-log_{2}(\frac{1}{2})} = 0 \quad \text{and} \quad \frac{\shannonentropy(\onecell[1,2,\pdataset^{rec1}])}{\shannonentropy(\onecell[1,2,\pdataset])} = \frac{-log_{2}(\frac{1}{3})}{-log_{2}(\frac{1}{3})} = 1.
\end{align*}
}
Conversely, we also have $\oldmetric(\pdataset^{rec2},\pdataset^{\origd}) = 0.5$ because:
{\small  
\begin{align*}\frac{\shannonentropy(\onecell[1,1,\pdataset^{rec2}])}{\shannonentropy(\onecell[1,1,\pdataset])} =  \frac{-log_{2}(\frac{1}{2})}{-log_{2}(\frac{1}{2})} = 1 ~ \text{and} ~ \frac{\shannonentropy(\onecell[1,2,\pdataset^{rec2}])}{\shannonentropy(\onecell[1,2,\pdataset])} = \frac{-log_{2}(1)}{-log_{2}(\frac{1}{3})} = 0.
\end{align*}
}
However, out of $6$ possible reconstructions for $\anexample_1$ (without any knowledge), $3$ are possible within $\pdataset^{rec1}$ while only $2$ are possible with $\pdataset^{rec2}$. 
Intuitively, $\pdataset^{rec2}$ yields more information (or, conversely, less uncertainty) than $\pdataset^{rec1}$, but $\oldmetric$ cannot account for this difference due to normalization and individual measure of entropy across examples' attributes. 
For notation consistency, we associate to these datasets their generalized counterparts $\generalizedpdataset^{Orig}$, $\generalizedpdataset^{rec1}$ and $\generalizedpdataset^{rec2}$, containing the exact same information (recall that probabilistic datasets are simply a particular case of generalized probabilistic datasets, in which the dataset's variables are statistically independent and uniformly distributed). 
Using our generalized metric introduced in Definition~\ref{def:distance_metric_generalized}, we have:
{\small
\begin{align*}
    \newmetric(\generalizedpdataset^{rec1},\generalizedpdataset^{\origd}) &=  \frac{\shannonentropy\left(\{\onecell[1,1,\generalizedpdataset^{rec1}], \onecell[1,2,\generalizedpdataset^{rec1}]\}\right)}{\shannonentropy\left(\{\onecell[1,1,\generalizedpdataset], \onecell[1,2,\generalizedpdataset]\}\right)}\\ 
    &= \frac{-log_{2}(\frac{1}{3})}{-log_{2}(\frac{1}{6})} \approx 0.613\\
    \text{and} \quad \newmetric(\generalizedpdataset^{rec2},\generalizedpdataset^{\origd}) &= \frac{\shannonentropy\left(\{\onecell[1,1,\generalizedpdataset^{rec2}], \onecell[1,2,\generalizedpdataset^{rec2}]\}\right)}{\shannonentropy\left(\{\onecell[1,1,\generalizedpdataset], \onecell[1,2,\generalizedpdataset]\}\right)}\\
    &= \frac{-log_{2}(\frac{1}{2})}{-log_{2}(\frac{1}{6})} \approx 0.387
\end{align*}
}
As lower values indicate less uncertainty (\emph{i.e.}, better reconstruction performances), we observe that $\newmetric$ successfully distinguishes between $\generalizedpdataset^{rec1}$ and $\generalizedpdataset^{rec2}$. 
Thus by avoiding the drawbacks of the normalization across dataset cells, the new metric $\newmetric$ successfully takes into account the specificities of the two probabilistic datasets.

\section{Generalized Probabilistic Datasets Reconstruction from Interpretable Models}
\label{sec:reconstruction_in_practice}

We now investigate how to quantify the success of a probabilistic reconstruction attack in practice.
First, we discuss how the attack success computation can be decomposed under reasonable assumptions regarding the structure of the interpretable model considered.
Then, we show how it can be computed without explicitly building the entire set of possible worlds, as long as one is able to count them.
Finally, we demonstrate that such simplification is possible for decision trees as well as rule lists models, and theoretically compare the reconstruction quality from these two hypothesis classes.

\subsection{General Case}

Let $\generalizedpdataset^{\interpretablemodel}$ be a generalized probabilistic dataset reconstructed from an interpretable model $\interpretablemodel$. 
As stated in Definition~\ref{def:distance_metric_generalized}, the success of the probabilistic reconstruction attack can be quantified using $\newmetric$. 
One can observe that the denominator $\left( \sum_{\exampleindex=1}^{\nsamples}{\sum_{\attributeindex=1}^{\nfeatures}{\shannonentropy(\onecell[\exampleindex,\attributeindex,\generalizedpdataset])}}\right)$ is a constant, only depending on the attributes' domains $\singledomain_{\attributeindex \in [1..\nfeatures]}$. 
Indeed, variables $\onecell[\exampleindex \in [1..\nsamples],\attributeindex,\generalizedpdataset]$ are uniformly distributed over $\singledomain_\attributeindex$ (the domain of attribute $\singleattribute_\attributeindex$) and so $\shannonentropy(\onecell[\exampleindex,\attributeindex,\generalizedpdataset]) = -log_{2}\left(\frac{1}{\lvert \singledomain_\attributeindex \rvert}\right)$. 
Thus:
\begin{align}
    \sum_{\exampleindex=1}^{\nsamples}{\sum_{\attributeindex=1}^{\nfeatures}{\shannonentropy(\onecell[\exampleindex,\attributeindex,\generalizedpdataset])}} = \nsamples \cdot {\sum_{\attributeindex=1}^{\nfeatures}{-log_{2}\left(\frac{1}{\lvert \singledomain_\attributeindex \rvert}\right)}} 
    \label{eq:denominator_value}
\end{align}

As the denominator in Equation~(\ref{eq:generalized_metric_line_1}) is a constant that can be easily computed, we will focus only on the numerator in the remaining of this section, using the following notation:
\begin{align}
\newmetric(\generalizedpdataset^{\interpretablemodel},\generalizedpdataset^{\origd}) \propto \shannonentropy\left(\{ \onecell[\exampleindex \in [1..\nsamples],\attributeindex \in [1..\nfeatures],\generalizedpdataset^{\interpretablemodel}]\}\right) \label{eq:proportional_numerator}
\end{align}

\subsubsection{Independence assumptions: decomposing the attack success computation.}

In the general case, the computation of the joint entropy of the generalized probabilistic dataset's cells must be done through its set of possible worlds $\possibleworlds(\generalizedpdataset{}^{\interpretablemodel})$, as shown in Equation~(\ref{eq:generalized_metric_line_2}). 
However, if one can establish the statistical independence of some of the $\onecell[\exampleindex,\attributeindex,\generalizedpdataset^{\interpretablemodel}]$ variables, this computation can be further decomposed. 
Indeed, the joint entropy of a set of statistically independent variables is equal to the sum of their individual entropies.
For instance, if the knowledge of model $\interpretablemodel$ applies to each example $\anexample_{\exampleindex \in [1..\nsamples]}$ independently, the sets of variables $\{\onecell[\exampleindex,\attributeindex \in [1..\nfeatures],\generalizedpdataset^{\interpretablemodel}]\}_{\exampleindex \in [1..\nsamples]}$ are independent from each other. 
This condition is satisfied if $\interpretablemodel$ is a decision tree or a rule list, because each example is captured by exactly one ``decision path'' (\emph{i.e.}, branch or rule). 
Indeed, this decision path reduces the set of possible reconstructions for each example $\anexample_\exampleindex$ independently from the other examples. 
By a slight abuse of notation, we let $\possibleworlds_\exampleindex(\generalizedpdataset^{\interpretablemodel})$ denote the set of possible worlds (\emph{i.e.}, reconstructions) for example (row) $\anexample_\exampleindex$. 
As a consequence, we have:
\begin{align}
    \newmetric(&\generalizedpdataset^{\interpretablemodel},\generalizedpdataset^{\origd}) \propto \sum_{\exampleindex=1}^{\nsamples}{\shannonentropy\left(\{ \onecell[\exampleindex,\attributeindex \in [1..\nfeatures],\generalizedpdataset^{\interpretablemodel}]\}\right)} \label{eq:proportional_numerator_decomposition_1}\\
    &\propto \sum_{\exampleindex=1}^{\nsamples}{\left(\sum_{\world_\exampleindex \in \possibleworlds_\exampleindex(\generalizedpdataset^{\interpretablemodel})}{-\mathbb{P}(\world_\exampleindex) \cdot log_{2}(\mathbb{P}(\world_\exampleindex))}\right)} \label{eq:proportional_numerator_decomposition_2}
\end{align}

While Equation~(\ref{eq:proportional_numerator_decomposition_2}) holds for both rule lists and decision trees, its computation can be further decomposed for the later. 
Indeed, in a decision tree each example is classified by exactly one branch, and such branch defines a conjunction of Boolean conditions over attributes' values, called \emph{splits}. 
Such conditions must all be satisfied for the example to be captured by the branch - hence all the concerned attributes' domains can be reduced individually.
As a consequence, this implies that all variables $\onecell[\exampleindex \in [1..\nsamples],\attributeindex \in [1..\nfeatures],\generalizedpdataset^{\interpretablemodel}]$ are actually statistically independent resulting in:
\begin{align}
    \newmetric(\generalizedpdataset^{\interpretablemodel},\generalizedpdataset^{\origd}) &\propto \sum_{\exampleindex=1}^{\nsamples}{\sum_{\attributeindex=1}^{\nfeatures}{\shannonentropy\left( \onecell[\exampleindex,\attributeindex,\generalizedpdataset^{\interpretablemodel}]\right)}} \label{eq:proportional_numerator_decomposition_3}
\end{align} 

Note that Equation~(\ref{eq:proportional_numerator_decomposition_3}) corresponds to the particular case studied in~\cite{DBLP:conf/dbsec/GambsGH12}, with the computation being exactly as for their proposed $\oldmetric$ metric (Definition~\ref{def:distance_metric}), with the only difference being the absence of normalization.
Observe that Equation~(\ref{eq:proportional_numerator_decomposition_3}) does not hold in general for rule list models due to the fact that for a given example, the information that it did not match previous rules within the rule list corresponds to negating a conjunction, hence producing a disjunction.
As a result, this potentially breaks the statistical independence between some of the $\{\onecell[\exampleindex,\attributeindex \in [1..\nfeatures],\generalizedpdataset^{\interpretablemodel}]\}$ variables.

\subsubsection{Uniform distribution assumptions: efficient attack success computation.}

The explicit enumeration of the possible worlds $\possibleworlds(\generalizedpdataset{}^{\interpretablemodel})$ is not practically conceivable for real-size datasets.
However, quantifying a probabilistic reconstruction attack success can sometimes be done only by computing their number $\lvert \possibleworlds(\generalizedpdataset{}^{\interpretablemodel}) \rvert$. 
Indeed, assuming a uniform probability distribution between them, one can then easily quantify the amount of uncertainty using $\newmetric$ (Definition~\ref{def:distance_metric_generalized}), as $\forall \world \in \possibleworlds(\generalizedpdataset{}^{\interpretablemodel}), \mathbb{P}(\world) = \frac{1}{\lvert \possibleworlds(\generalizedpdataset{}^{\interpretablemodel}) \rvert}$, resulting in:
\begin{align}
\sum_{\world \in \possibleworlds(\generalizedpdataset^{\interpretablemodel})}{- \mathbb{P}(\world) \cdot log_{2}(\mathbb{P}(\world))} = -log_{2}\left(\frac{1}{\lvert \possibleworlds(\generalizedpdataset{}^{\interpretablemodel}) \rvert}\right) \label{eq:computing_section_general_line1}
\end{align}
Remark that only the number of possible worlds $\lvert \possibleworlds(\generalizedpdataset{}^{\interpretablemodel}) \rvert$ is needed to compute Equation~(\ref{eq:computing_section_general_line1}). 
In the general case, this number cannot be retrieved without building $\possibleworlds(\generalizedpdataset{}^{\interpretablemodel})$ explicitly.
However, several types of interpretable models enable to compute $\lvert \possibleworlds(\generalizedpdataset{}^{\interpretablemodel}) \rvert$ efficiently (\emph{i.e.}, without building $\possibleworlds(\generalizedpdataset{}^{\interpretablemodel})$). 
For instance, this is the case when reconstructing generalized probabilistic datasets from decision tree or rule list models.
Plugging together Equations~(\ref{eq:proportional_numerator_decomposition_2}) and~(\ref{eq:computing_section_general_line1}), we have:
\begin{align}
    \newmetric(\generalizedpdataset^{\interpretablemodel},\generalizedpdataset^{\origd}) &\propto \sum_{\exampleindex=1}^{\nsamples}{-log_{2}\left(\frac{1}{\lvert \possibleworlds_\exampleindex(\generalizedpdataset{}^{\interpretablemodel}) \rvert}\right)} \label{eq:computing_section_general_line2}
\end{align}

In the next subsections, we demonstrate how $\lvert \possibleworlds_\exampleindex(\generalizedpdataset{}^{\interpretablemodel}) \rvert _ {\exampleindex \in [1..\nsamples]}$ can be computed in polynomial time (with respect to the model's size) for decision trees and rule lists. 
Note that the assumptions performed in this subsection are realistic. 
In particular, the uniform distribution assumption simply means that if the model at hand is compatible with several reconstructions (deterministic datasets), they are all equally likely (one can not state whether one is more likely than the others). Furthermore, the independence assumption (between the dataset rows, as in~(\ref{eq:proportional_numerator_decomposition_1})) is mainly challenging when dealing with ensemble models (and matching the knowledge of the different base learners is an additional challenge).

\subsection{Decision Trees}


Let $\decisiontree$ be a decision tree with $\nbbranches$ branches, in which each branch $\branch_{\branchindex \in [1..\nbbranches]}$ is a conjunction of Boolean assertions over attributes' values ending with a leaf prediction. 
The value $\numb(\branch_\branchindex)$ represents the number of different examples (\emph{i.e.}, number of different combinations of attributes values) that satisfy $\branch_\branchindex$.
It can be computed by multiplying the cardinalities of the reduced domains.
Thus, for each example $\anexample_\exampleindex$ classified by branch $\branch_\branchindex$, we have $\lvert \possibleworlds_\exampleindex(\generalizedpdataset^{\decisiontree}) \rvert = \numb(\branch_\branchindex)$.
Additionally, $\nbexclassified_{\branchindex \in [1..\nbbranches]}$ is defined as the support of each leaf (\emph{i.e.}, the number of training examples captured by the leaf, as indicated in the decision tree of Figure~\ref{fig:example_toy_dt}).
Importantly, the tree branches partition the set of examples (as the leaves' supports are all disjoints), so we have $\sum_{\branchindex \in [1..\nbbranches]}{\nbexclassified_{\branchindex}} = \nsamples$. 
Furthermore, the sum of Equation~(\ref{eq:computing_section_general_line2}) which was performed over all $\nsamples$ examples can be replaced with a sum over the $\nbbranches$ branches, with the entropy of each branch $\branch_\branchindex$ being weighted by its support $\nbexclassified_{\branchindex}$. 
Plugging these new notions into Equation~(\ref{eq:computing_section_general_line2}), we obtain that the overall joint entropy of the reconstructed probabilistic version of $\decisiontree$'s training set is:
\begin{align}
    \newmetric(\generalizedpdataset^{\decisiontree},\generalizedpdataset^{\origd}) &\propto  \sum\limits_{\branchindex=1}^{\nbbranches}{- \nbexclassified_\branchindex \cdot log_2\left(\frac{1}{\numb(\branch_\branchindex)}\right)} \label{eq:dt_dist_efficient_computation}
\end{align}


\subsection{Rule Lists} 
\label{subsection:rule_lists_metric}

Let $\rulelist = \arule[1] \ldots \arule[\rulelistlength]$ be a rule list, following the notation introduced in~\cite{DBLP:journals/ml/Rivest87}. 
Each term $\antecedent_{\ruleindex \in [1..\rulelistlength]}$ is a conjunction of Boolean assertions over attributes' values and $\consequent_{\ruleindex \in [1..\rulelistlength]}$ is a prediction. 
Rule $\rulelistlength$ is the default decision, with $\antecedent_{\rulelistlength}$ being the constant value \textsf{True}. 
Similarly to the leaves of a decision tree, each rule $\ruleindex$ is associated with its support $\nbexclassified_{\ruleindex}$.
Again, let $\numb(\antecedent_\ruleindex)$ denote the number of different examples (\emph{i.e.}, number of different combinations of attributes values) that satisfy $\antecedent_\ruleindex$. 
As a branch, a rule corresponds to a conjunction, hence $\numb(\antecedent_\ruleindex)$ can be computed easily by simply multiplying the cardinalities of the attributes' reduced domains.

Finally, we define $\forall 1 \leq \ruleindex \leq \rulelistlength,~ \capt[\rulelist](\antecedent_\ruleindex)$ as the number of possible different examples (\emph{i.e.}, number of different combinations of attributes values) that $\antecedent_\ruleindex$ captures \emph{within $\rulelist$} (\emph{i.e.}, examples satisfying $\antecedent_\ruleindex$ while not matching the antecedents of the previous rules within $\rulelist$). 
As a particular case, note that we always have $\capt[\rulelist](\antecedent_1) = \numb(\antecedent_1)$ as the first rule of any rule list is always applied first. 
For $1 \leq \ruleindex \leq \rulelistlength$, a straightforward general formulation is:
\begin{align}
\capt[\rulelist](\antecedent_\ruleindex) = \numb(\antecedent_\ruleindex \land \bigwedge_{\ruleindexbis=1}^{\ruleindex-1}{\neg \antecedent_\ruleindexbis})
\end{align}

The main challenge is that $\numb(\bigwedge_{\ruleindexbis=1}^{\ruleindex-1}{\neg \antecedent_\ruleindexbis})$, in which $\bigwedge_{\ruleindexbis=1}^{\ruleindex-1}{\neg \antecedent_\ruleindexbis}$ is the conjunction of the negations of the previous rules' antecedents, cannot be computed directly as $\antecedent_{\ruleindexbis\in[1..\ruleindex-1]}$ may overlap. 
Indeed, each antecedent $\antecedent_{\ruleindexbis}$ is a conjunction - hence its negation is a disjunction. 
More precisely, overall we get a conjunction of disjunctions, which means that the number of possible examples it characterizes cannot be computed by simply multiplying attributes' cardinalities as the different disjunctions may overlap.
By a slight abuse of notation, we define for $1 \leq \ruleindexbis \leq \ruleindex \leq \rulelistlength,~ \capt[\rulelist](\antecedent_\ruleindexbis,\antecedent_\ruleindex)$ as the number of possible different examples (\emph{i.e.}, number of different combinations of features values) that $\antecedent_\ruleindex$ could capture but that are actually captured by $\antecedent_\ruleindexbis$ in $\rulelist$:
\begin{align}
    \capt[\rulelist](\antecedent_\ruleindexbis,\antecedent_\ruleindex) = \numb(\antecedent_\ruleindexbis \land \antecedent_\ruleindex) - \sum\limits_{\ruleindexter=1}^{\ruleindexbis-1}{\capt[\rulelist](\antecedent_\ruleindexter,(\antecedent_\ruleindexbis \land \antecedent_\ruleindex))}
\end{align}
The first term corresponds to the overlap between $\antecedent_\ruleindexbis$ and $\antecedent_\ruleindex$, while the second one subtracts the unique examples within this overlap that are actually captured by rules placed before $\antecedent_\ruleindexbis$ in $\rulelist$.
Then:
\begin{align}
\capt[\rulelist](\antecedent_\ruleindex) &= \capt[\rulelist](\antecedent_\ruleindex,\antecedent_\ruleindex) \\ &= \numb(\antecedent_\ruleindex) - \sum\limits_{\ruleindexbis=1}^{\ruleindex-1}{\capt[\rulelist](\antecedent_\ruleindexbis,\antecedent_\ruleindex)} \label{eq:rule_list_recursive_efficient_computation}
\end{align}

Just like the branches of a decision tree, the rules within a rule list partition the set of examples (as each example is captured by exactly one rule in the rule list). Then, the sum over all $\nsamples$ examples in Equation~(\ref{eq:computing_section_general_line2}) can be reformulated using a sum over the $\rulelistlength$ rules, with each rule's entropy being weighted by its support.
Then, plugging~(\ref{eq:rule_list_recursive_efficient_computation}) into~(\ref{eq:computing_section_general_line2}), we obtain:
\begin{align}
    &\newmetric(\generalizedpdataset^{\rulelist},\generalizedpdataset^{\origd}) \nonumber \\
    &\propto \sum\limits_{\ruleindex=1}^{\rulelistlength}{- \nbexclassified_\ruleindex \cdot  log_{2}\left(\frac{1}{\numb(\antecedent_\ruleindex) - \sum\limits_{\ruleindexbis=1}^{\ruleindex-1}{\capt[\rulelist](\antecedent_\ruleindexbis,\antecedent_\ruleindex)}}\right)} \label{eq:rl_dist_efficient_computation}
\end{align}

\paragraph*{Comparing Decision Trees and Rule Lists} Comparing~(\ref{eq:rl_dist_efficient_computation}) to~(\ref{eq:dt_dist_efficient_computation}), we observe that an additional term is subtracted to the denominator of~(\ref{eq:rl_dist_efficient_computation}). 
This term corresponds to the information that the examples captured by rule $\ruleindex$ did not match any of the previous rules $\ruleindexbis < \ruleindex$ within $\rulelist$.
By lowering the denominator, it raises the overall success of the probabilistic reconstruction attack.
There is no such term in~(\ref{eq:dt_dist_efficient_computation}) because there can be no overlap between a decision tree's leaves' supports.
On the contrary, the rules within a rule list can overlap because they are ordered.
Overall, these theoretical results confirm that rule lists are more expressive than decision trees, encoding more information than a decision tree of equivalent size~\cite{DBLP:journals/ml/Rivest87}. \revision{The reconstruction methods depicted in Sections~\ref{sec:probabilistic_reconstruction_gambs} and~\ref{sec:motivation_generalization} have a computational cost linear in the models' sizes, as they simply follow each decision path once. However, the computation is more expensive for rule lists if one wants to avoid explicitly building the set of possible reconstructions, and the computational cost of~(\ref{eq:rl_dist_efficient_computation}) is rather factorial in the number of rules (which can be proved by induction). This suggests that the additional information brought by rule lists is also more challenging to extract.}



\section{Experiments}
\label{sect_experiments}

While our proposed metric precisely quantifies the amount of information an interpretable model carries regarding its training dataset, the aim of this section is to illustrate its practical usefulness through an example use.
More precisely, we will investigate the differences between optimal and heuristically-built models, for both rule lists and decision trees.

\subsection{Setup}\label{subsec:setup}

In these experiments, we use both optimal and heuristic learning algorithms to compute decision trees and rule lists of varied sizes.
Furthermore, optimal models are learnt optimizing solely accuracy, to avoid interference with other regularization terms.
All details regarding the considered experimental setup are provided hereafter.

\paragraph{Learning algorithms} We use the following learning algorithms:
\begin{itemize}
    \item \textbf{Optimal decision trees.} We use the \exactdt{} algorithm~\cite{DBLP:conf/aaai/AglinNS20,DBLP:conf/ijcai/AglinNS20} through its Python binding\footnote{\url{https://github.com/aia-uclouvain/pydl8.5}}.
    \item \textbf{Heuristic decision trees.} We use an optimized version of the \cart{} greedy algorithm~\cite{DBLP:books/wa/BreimanFOS84}, as implemented within the \sklearn{}\footnote{\url{https://scikit-learn.org/}} Python library~\cite{scikit-learn} with its \DecisionTreeClassifier{} object. We coin this method \heuristicdt{}.
    \item \textbf{Optimal rule lists.} We use the \corels{} algorithm~\cite{angelino2017learning,angelino2018learning} through its Python binding\footnote{\url{https://github.com/corels/pycorels}}.
    \item \textbf{Heuristic rule lists.} While some implementations exist in the literature for building \emph{heuristic rule lists} (for example, one is provided within the \texttt{imodels}\footnote{\url{https://github.com/csinva/imodels}} library\footnote{\texttt{imodels} is a Python library gathering tools to learn different types of popular interpretable machine learning models.}~\cite{imodels2021}), they do not offer precise control over the desired rule support and/or maximum rule list depth. 
For this reason, we implemented a \cart{}-like greedy algorithm (close to the \texttt{imodels}' implementation), that we coin \greedyRL{}.
In a nutshell, this algorithm selects the rule yielding to the best Gini impurity improvement at each level of the rule list, in a top-down manner.
\end{itemize}

\paragraph{Datasets} We use two datasets (binarized, as required by~\corels{}) which are very popular in the trustworthy machine learning literature.
First, the UCI Adult Income dataset\footnote{\url{https://archive.ics.uci.edu/ml/datasets/adult}}~\cite{Dua:2019} contains data regarding the 1994 U.S. census, with the objective of predicting whether a person earns more than \$50K/year.
Numerical features are discretized using quantiles and categorical features are one-hot encoded. 
The resulting dataset includes $48,842$ examples and $24$ binary features. 
As \exactdt{} was unable to learn optimal models within the specified time and memory limits for the largest size constraints, we randomly sub-sample $10\%$ of the whole dataset.
Second, the COMPAS dataset (analyzed by~\cite{angwin2016machine}) gathers records about criminal offenders in the Broward County of Florida collected from 2013 and 2014, with the task being recidivism prediction. 
We consider its discretized  version  used  to evaluate \corels{} in \cite{angelino2017learning}, consisting in $7,214$ examples characterized with $27$ binary features\footnote{\url{https://github.com/corels/pycorels/blob/master/examples/data/compas.csv}}.


\paragraph{Experimental Parameters}  For each experiment, we randomly select $80\%$ of the dataset to form a training set, and use the remaining $20\%$ as a test set to ensure that models generalize well. 
We repeat the experiment five times using different seeds for the random train/test split, and report results averaged across the five runs.
All experiments are run on a computing cluster over a set of homogeneous nodes using Intel Platinum 8260 Cascade Lake @ 2.4Ghz CPU.
Each training phase is limited to one hour of CPU time and 12 GB of RAM.
Within the proposed experimental setup, all models produced by the optimal learning algorithms (\exactdt{} for decision trees or \corels{} for rule lists) are certifiably optimal.

\paragraph{Models Learning} We set various constraints to the decision tree building algorithms, using maximum tree depths 
between $1$ and $10$ (ranging linearly by steps of $1$) and (relative) minimum leaf supports 
between $0.01$ and $0.05$ (ranging linearly by steps of $0.01$).
For the rule list learning algorithms, we proceed identically and generate rule lists with various size constraints, using maximum depths \emph{(number of rules within the rule list)} between $1$ and $10$ (ranging linearly by steps of $1$) and (relative) minimum rule supports 
between $0.01$ and $0.05$ (ranging linearly by steps of $0.01$). 
As we are interested in the optimality guarantee, we consider rules consisting in a single binary attribute (or its negation). 
Indeed, in our experiments, \corels{} was unable to reach and certify optimality while also considering conjunction of features, as it dramatically increases the number of rules - and consequently, the algorithm search space.
Finally, we set \corels{}'s sparsity regularization coefficient to a value small enough (\emph{i.e.}, smaller than $\frac{1}{\nsamples}$) to ensure that only accuracy is optimized.
All methods' parameters are left to their default value.

\paragraph{Resources} 
All scripts needed to reproduce our experiments, as well as our implementation of the greedy rule list learning algorithm \greedyRL{} are publicly available\footnote{https://github.com/ferryjul/ProbabilisticDatasetsReconstruction}. 

\subsection{Results} \label{sec:results}

\def\figsize{0.40}
\def\legsize{0.99}

\begin{figure*}[htb!]
     \begin{center}
    
    \begin{subfigure}{\textwidth}
        \centering
        
        \includegraphics[width=\figsize\textwidth]{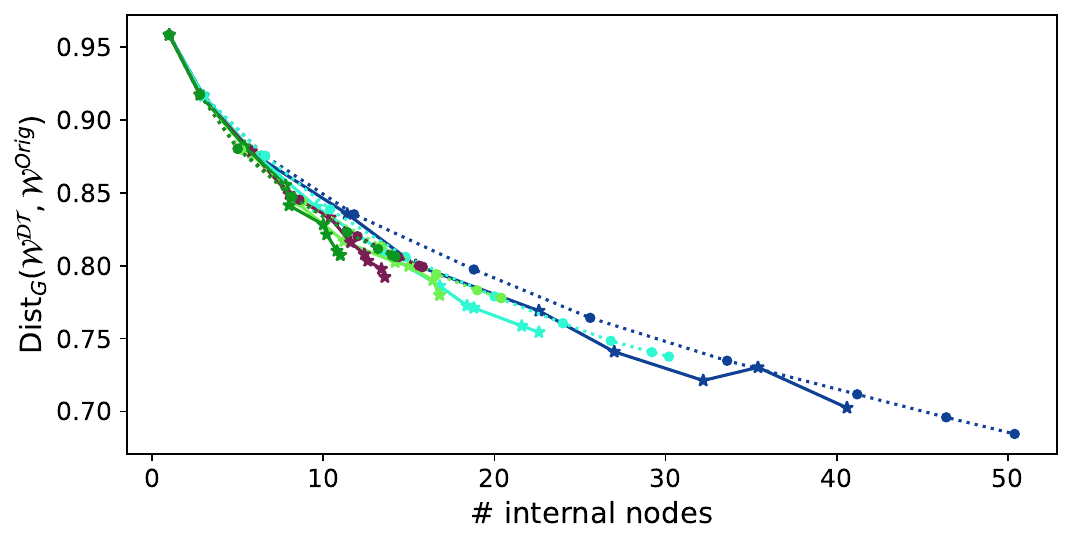}
        \includegraphics[width=\figsize\textwidth]{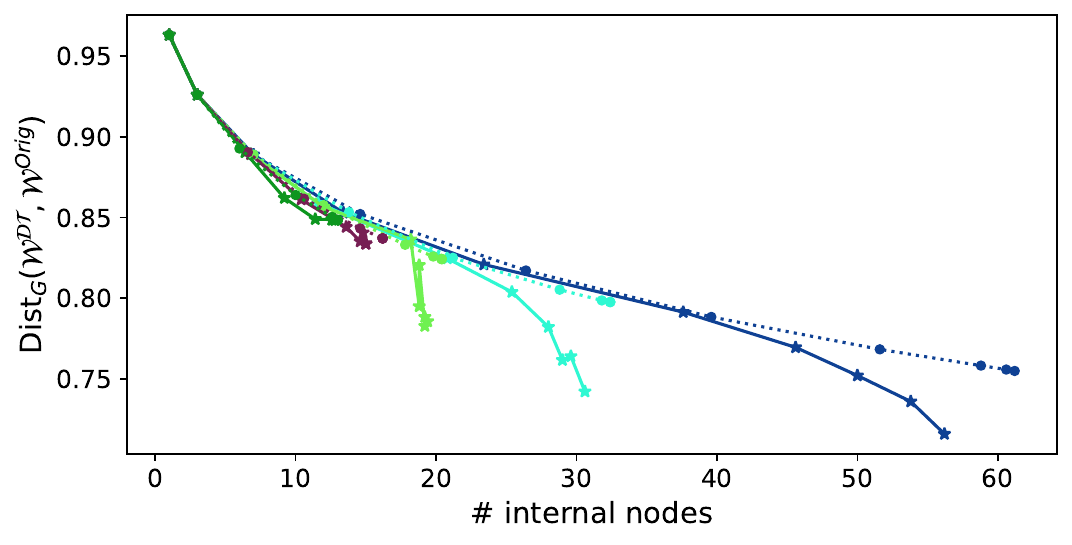}

        \caption{Entropy reduction as a function of the tree size (number of splits/internal nodes). Note that the number of internal nodes is restricted by the maximum depth constraint, but also by the enforced minimum leaf support (as mentioned in section~\ref{subsec:setup}), which explains why the largest trees only have slightly more than $60$ nodes. \label{fig:results_trees_entropy_reduction_f_n_elementary_tokens}}
        
    \end{subfigure}
    
    \par\bigskip 
        \begin{subfigure}{\textwidth}
        \centering
        
        \includegraphics[width=\figsize\textwidth]{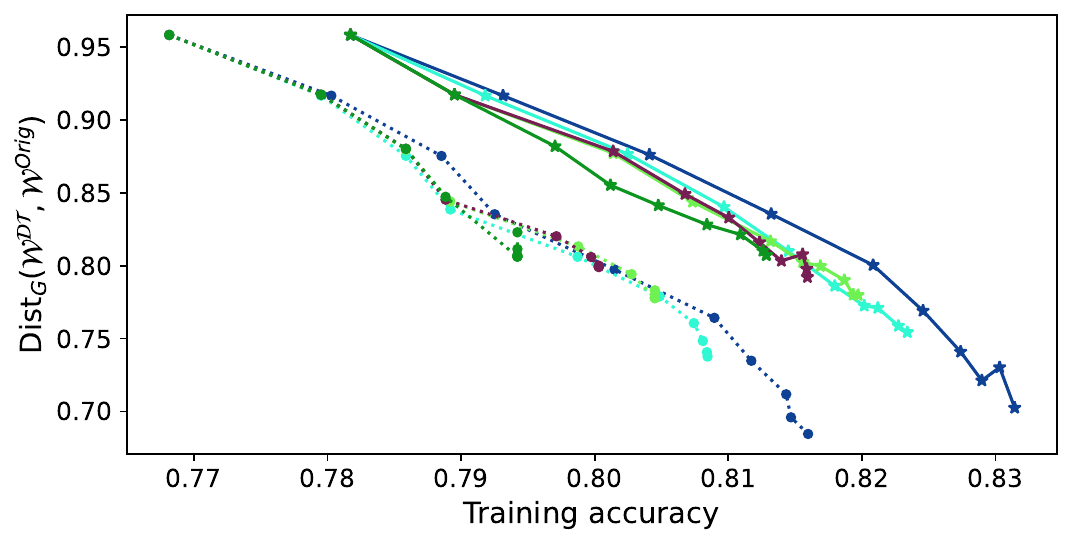}
        \includegraphics[width=\figsize\textwidth]{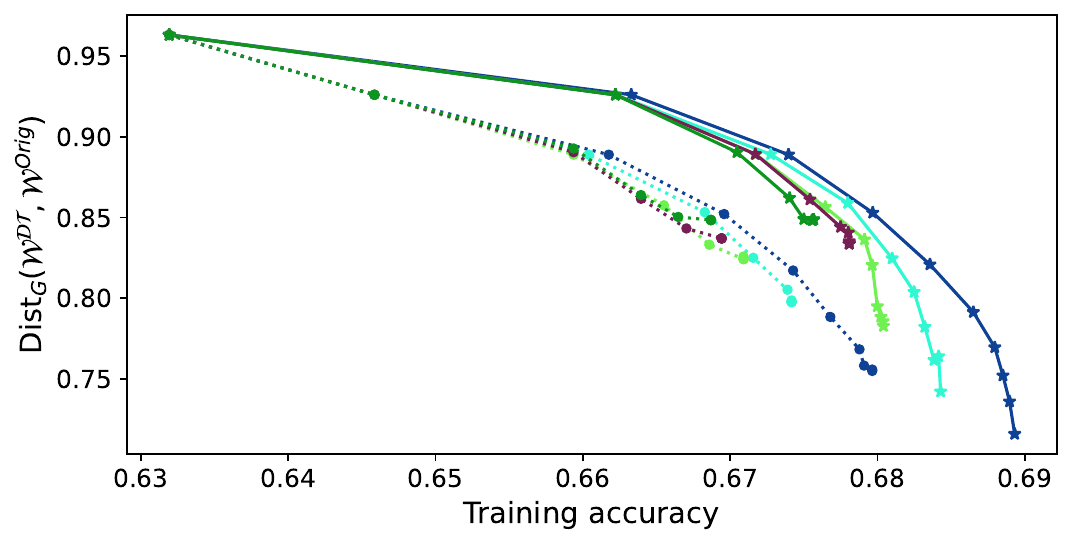}
      
        \caption{Entropy reduction as a function of training accuracy.~\label{fig:results_trees_entropy_reduction_f_training_acc}}
        
    \end{subfigure}

     \vskip 10 pt

    \includegraphics[width=\legsize\textwidth]
   {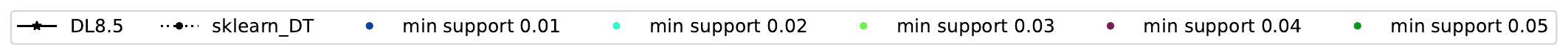}

    \caption{Results of our experiments comparing optimal and greedily-built decision trees (learnt respectively with \exactdt{} and \heuristicdt{}), for different (relative) minimum leaf support values. Left: Adult Income dataset, Right: COMPAS dataset.}  
     
    \label{fig:results_trees_all}
    \end{center}
\end{figure*}

\begin{figure*}[ht!]
     \begin{center}
    
    \begin{subfigure}{\textwidth}
        \centering
        
        \includegraphics[width=\figsize\textwidth]{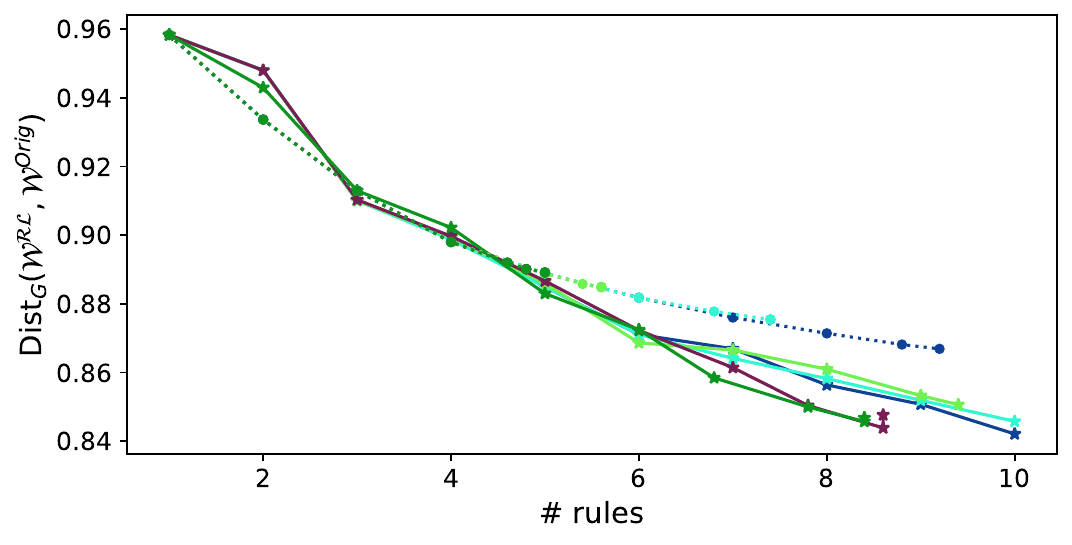}
        \includegraphics[width=\figsize\textwidth]{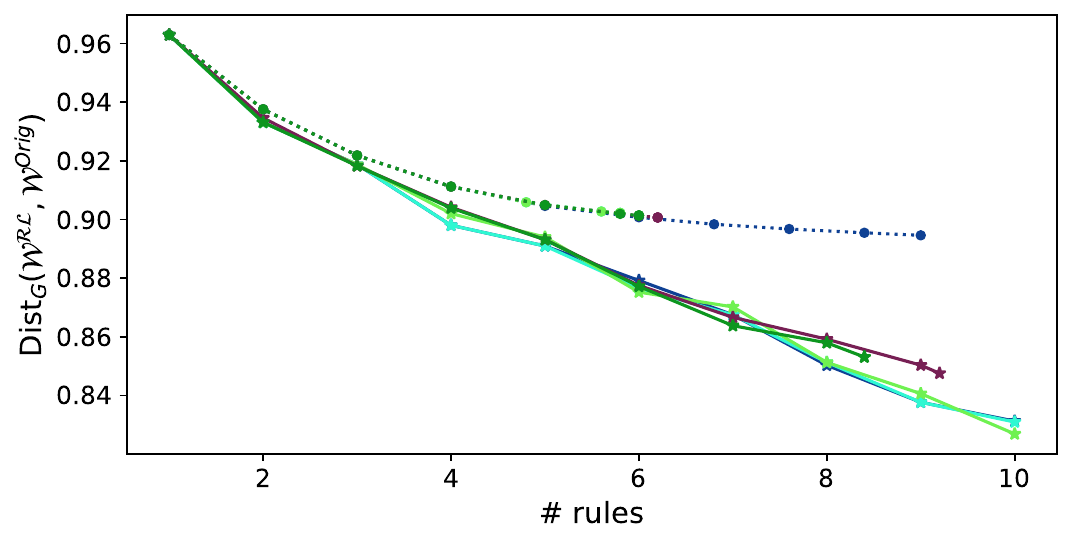}

        \caption{Entropy reduction as a function of the rule list size (number of rules).~\label{fig:results_rule_lists_entropy_reduction_f_n_elementary_tokens}}
        
    \end{subfigure}
    
    \par\bigskip 
        
    \begin{subfigure}{\textwidth}
        \centering
        
        \includegraphics[width=\figsize\textwidth]{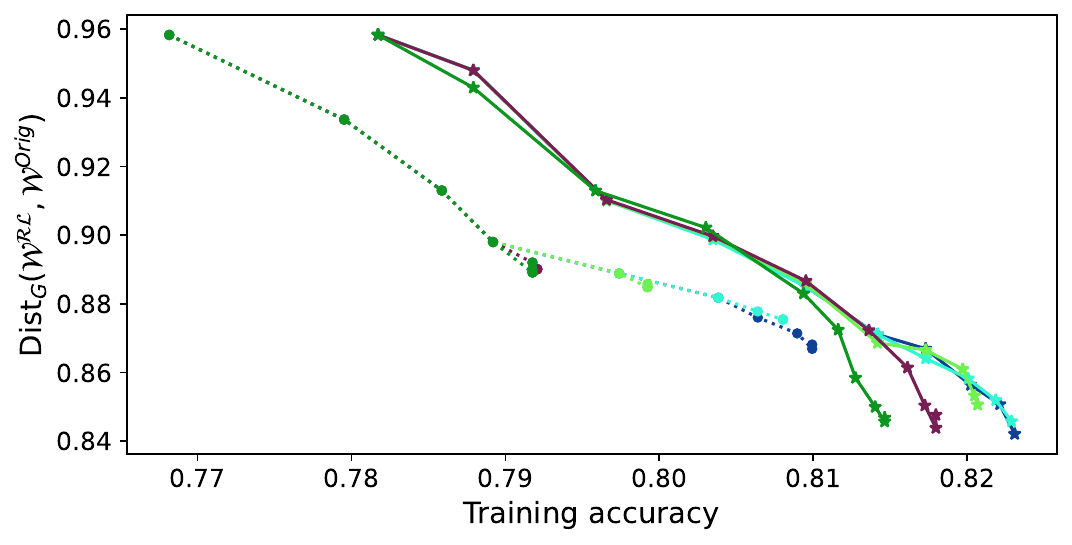}
        \includegraphics[width=\figsize\textwidth]{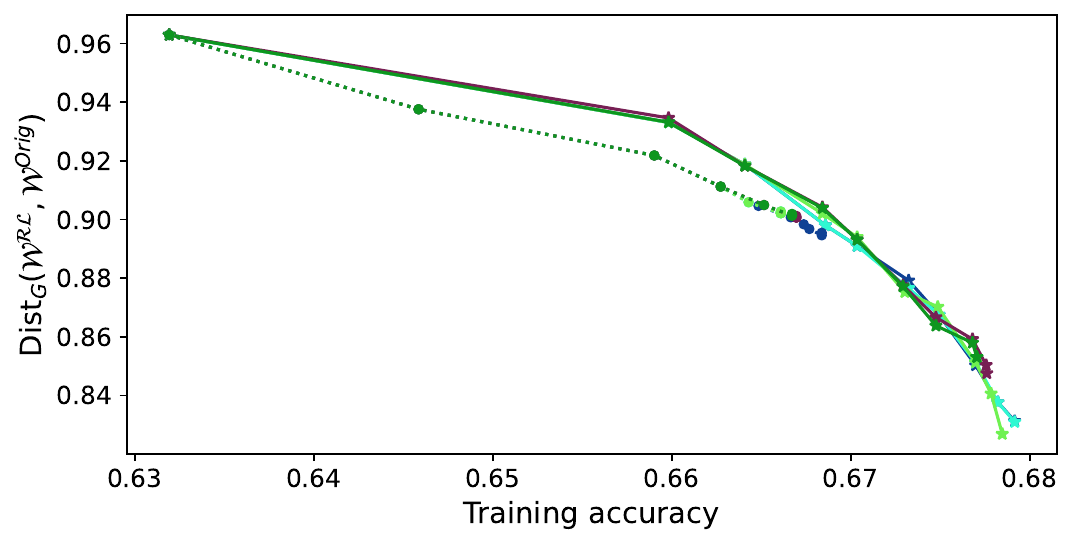}
      
        \caption{Entropy reduction as a function of training accuracy.~\label{fig:results_rule_lists_entropy_reduction_f_training_acc}}
        
    \end{subfigure}

     \vskip 10 pt

    \includegraphics[width=\legsize\textwidth]
   {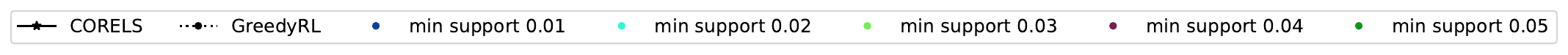}

    \caption{Results of our experiments comparing optimal and greedily-built rule lists (learnt respectively with \corels{} and \greedyRL{}), for different (relative) minimum rule support values. Left: Adult Income dataset, Right: COMPAS dataset.}    
     
    \label{fig:results_rule_lists_all}
    \end{center}
\end{figure*}

\def\localegsize{0.20}

\begin{figure*}[htb!]
    \begin{center}
   
        \begin{subfigure}{\textwidth}
            \centering
            
            \includegraphics[width=\figsize\textwidth]{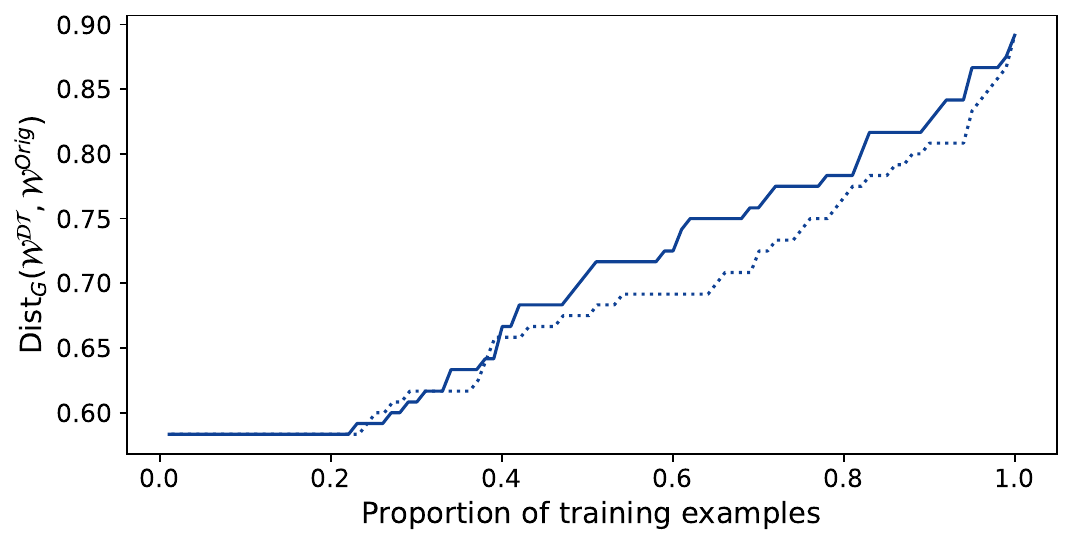}
            \includegraphics[width=\figsize\textwidth]{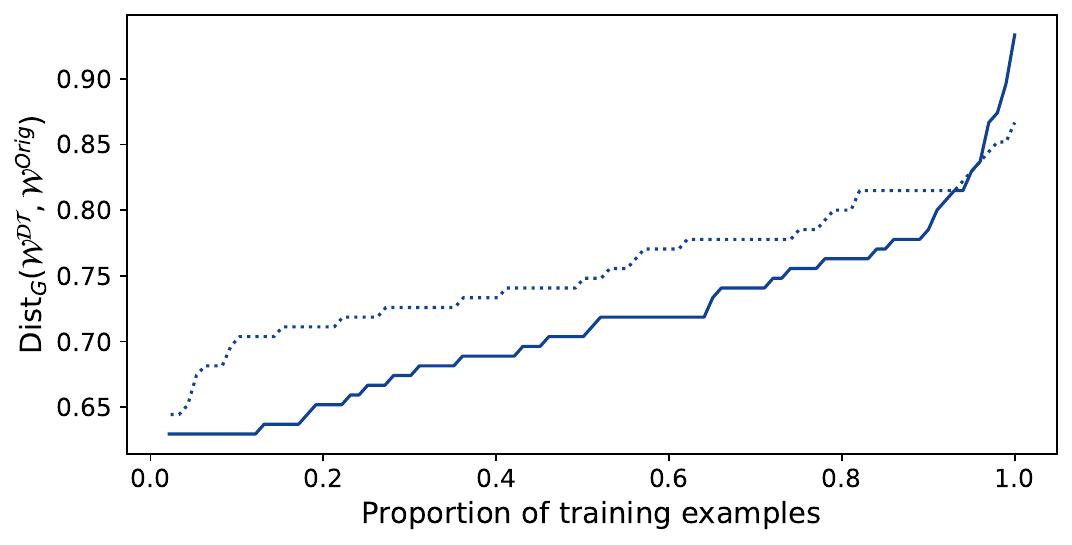}

           \includegraphics[width=\localegsize\textwidth]{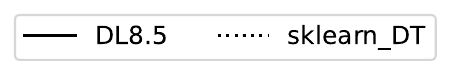}
           
            \caption{Optimal and greedily-built decision trees, learned respectively with the \exactdt{} and \heuristicdt{} algorithms.}~\label{fig:results_decision_trees_disparate_info_leak} 
            
        \end{subfigure}
 
        \par\bigskip 
    
        \begin{subfigure}{\textwidth}
            \centering
            
            \includegraphics[width=\figsize\textwidth]{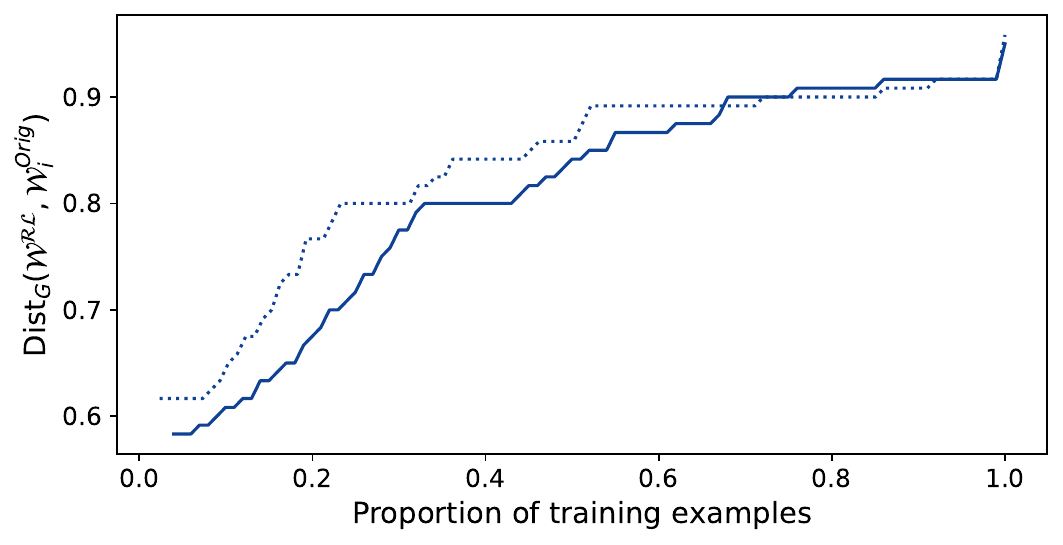}
            \includegraphics[width=\figsize\textwidth]{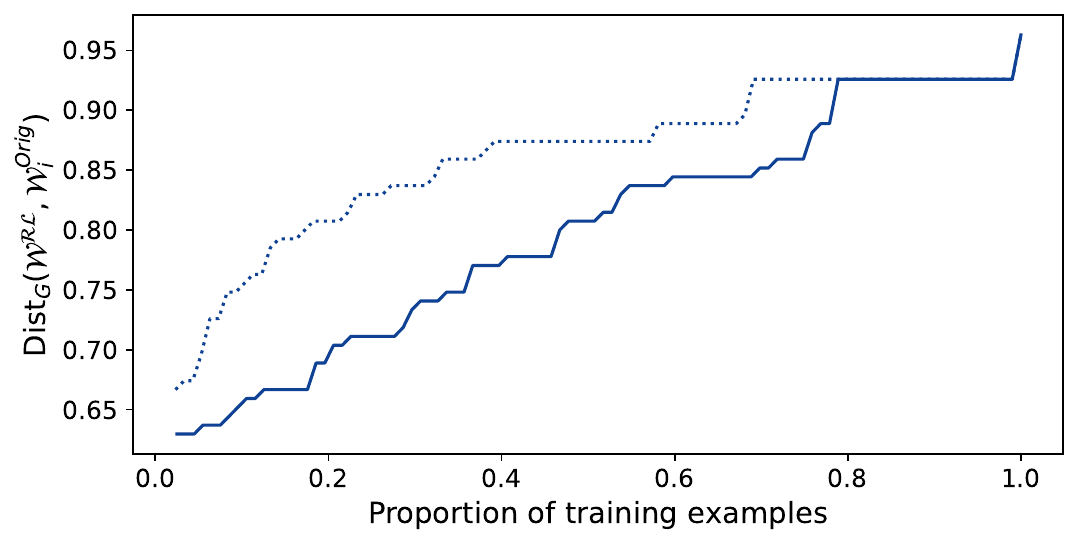}
            
           \includegraphics[width=\localegsize\textwidth]{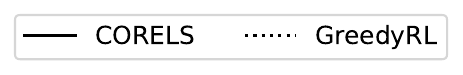}
        
        \caption{Optimal and greedily-built rule lists, learned respectively with the \corels{} and \greedyRL{} algorithms.}~\label{fig:results_rule_lists_disparate_info_leak} 

        \end{subfigure}
            
        \caption{Illustration of the \emph{disparate information leak} phenomenon, for both optimal and greedily-built decision trees and rule lists, learned with the largest considered size constraints, \emph{i.e.,} maximum depth $10$ and minimum (relative) support $0.01$. More precisely, we report the proportion of training examples 
    for which the entropy reduction ratio 
    is at most at a given value. Left: Adult Income dataset, Right: COMPAS dataset.}~\label{fig:results_disparate_info_leak} 
    
    \end{center}
\end{figure*}

After having learnt optimal and heuristic decision trees and rule lists under various constraints, we compute the amount of information they contain regarding their training sets using $\newmetric$, leveraging the computational tricks presented in Equations~(\ref{eq:dt_dist_efficient_computation}) and~(\ref{eq:rl_dist_efficient_computation}).
\revision{As discussed in Section~\ref{sec:reconstruction_in_practice}, we make the uniform distribution assumption, because the sole structure of a decision tree or rule list does not allow distinguishing between different compatible reconstructions. We also make the independence assumption between the different examples, because in both decision trees and rule lists, each example is captured by a single decision path (\emph{i.e.,} leaf or rule). However, we do not make the independence assumption between the different attributes of each example, because as illustrated in Section~\ref{sec:motivation_generalization}, it does not hold for rule lists models.}
Recall that lower uncertainty values indicate better reconstruction performances.
We relate this value to two dimensions: the sizes of the models and their training accuracy.
The former corresponds respectively to the number of splits performed in a decision tree or to the number of rules for a width-1 rule list.
The later indicates the model's performance on its training set - \emph{i.e.}, exactly what we aim at optimizing. 

Results are provided for our experiments comparing exact and greedily-built decision trees and rule lists respectively in Figures~\ref{fig:results_trees_all} and~\ref{fig:results_rule_lists_all}.
We observe the same trends for the two types of models.
First, one can observe in Figures~\ref{fig:results_trees_entropy_reduction_f_n_elementary_tokens} and \ref{fig:results_rule_lists_entropy_reduction_f_n_elementary_tokens} that optimal models usually represent more information in a more compact way: the reconstruction uncertainty 
decreases faster for optimal models than with greedily-built ones.
However, while for a given size optimal models contain more information regarding their training data, they are also way more accurate. 
This dimension is observed in Figures~\ref{fig:results_trees_entropy_reduction_f_training_acc} and \ref{fig:results_rule_lists_entropy_reduction_f_training_acc}. 
More precisely, we consistently observe that for a given accuracy level, optimal models always leak less information regarding their training data.
These observations can be explained by the nature of the learning algorithms. 
On the one side, greedy algorithms make heuristic choices iteratively. 
These choices are usually sub-optimal, and thus while leading to sub-optimal models (in terms of accuracy), they can also cause unnecessary leaks regarding their training data. 
On the other side, optimal learning algorithms perform global optimization and encode exactly the information needed in the most effective way.

For both datasets and types of models, the entropy reduction is not uniformly distributed across all training examples.
Indeed, we plot in Figure~\ref{fig:results_disparate_info_leak} the minimum entropy reduction ratio as a function of the proportion of concerned training examples.
One can observe that the amount of information contained by the learnt models varies significantly between different training examples.
For instance, in the experiments using optimal rule lists with maximum size $10$ and minimum support $0.01$ on the Adult Income dataset (Figure~\ref{fig:results_rule_lists_disparate_info_leak} (left)), the less exposed training examples have an entropy reduction ratio above $0.95$: knowledge of the rule list removes only $5$\% of the uncertainty regarding such samples. 
For the most exposed examples, this number becomes smaller than $0.60$: knowledge of the rule list removes more then $40$\% of the uncertainty regarding such samples.
This \emph{disparate information leak} is intuitive: an example classified by a very long branch of a tree goes through many nodes, which gives more information regarding its features. 
This phenomenon is observed in all our experiments, with roughly identical distribution of the uncertainty reduction over the training datasets. 
It suggests that, behind average-case uncertainty reduction as reported in Figures~\ref{fig:results_trees_all} and \ref{fig:results_rule_lists_all}, investigating per-example uncertainty reductions can also be insightful. 

One can note that averaging the curves of Figure~\ref{fig:results_disparate_info_leak} leads to the computation of dataset-wide metrics as shown in Figures~\ref{fig:results_trees_all} and \ref{fig:results_rule_lists_all}.
For instance, we observe in Figure~\ref{fig:results_rule_lists_disparate_info_leak} that for most proportions of training samples, rule lists learnt using \corels{} exhibit a lower entropy reduction ratio than those produced by \greedyRL{}.
As aforementioned, these experiments use the largest considered rule lists (learned with maximum depth 10 and minimum support 0.01) for both methods, corresponding to the rightmost points on Figure~\ref{fig:results_rule_lists_entropy_reduction_f_n_elementary_tokens}.
Observing these particular points, one can see that rule lists built with \corels{} indeed exhibit lower entropy reduction ratios than those built by \greedyRL{}, which is consistent with Figure~\ref{fig:results_rule_lists_disparate_info_leak}.

We observe a different trend for the decision trees learnt on the Adult Income dataset (Figure~\ref{fig:results_decision_trees_disparate_info_leak} (left)): the models built with the greedy \heuristicdt{} algorithm exhibit lower entropy reduction ratios than the optimal ones produced by \exactdt{}{}, hence containing more information. 
Again, these models correspond to the rightmost points on Figure~\ref{fig:results_trees_entropy_reduction_f_n_elementary_tokens} (left).
For these experiments, the optimal models learnt with the \exactdt{} algorithm with the largest size constraint are indeed more compact than those produced by \heuristicdt{} and contain less information overall.
As aforementioned, this illustrates a drawback of greedy learning algorithms: by performing local (possibly sub-optimal) choices, they can produce models performing non-necessary or redundant operations, leaking additional information regarding their training data. 
This dimension is further explored in the Appendix~\ref{appendix:additional_results_p_datasets}, where we relate the actual models' sizes and entropy reduction ratios to the constraints enforced during learning.
 
Finally, comparing decision trees and rule lists empirically as was done theoretically in Section~\ref{subsection:rule_lists_metric} could also be insightful.
In particular, one could assess whether the rules' ordering, which allows the rules within a rule list to overlap (while the branches of a decision tree are all disjoint), empirically provides more information regarding the training data as was expected theoretically.
However, such an experiment requires learning optimal rule lists whose rules' widths (\emph{i.e.}, number of attributes involved in a rule's conjunction) match the depth of the tree's branches, which is computationally challenging. Indeed, considering sub-optimal models would bias the comparison as the results would depend on the performances of the learning algorithms rather than those of the models themselves.

\section{Conclusion}

We extended previous work and proposed generic tools to represent and precisely quantify the amount of information an interpretable model encodes regarding its training data. 
Such tools, and in particular the proposed generalized probabilistic datasets and the metric quantifying their amount of uncertainty, are generic enough to encode any type of knowledge - and hence are suitable to model a reconstructed dataset from any type of interpretable models.
While their practical use may be computationally challenging in the general case, we demonstrated theoretically that they can be employed efficiently under reasonable assumptions. 
Furthermore, we empirically illustrated their usefulness 
through an example use case: assessing the effect of optimality in training machine learning models. 
Designing appropriate data structures to efficiently represent our generalized probabilistic datasets in the general case and handle more complex types of models such as random forests or other ensemble classifiers is an important challenge.
However, this is out of the scope of this paper - as we restrict our attention to interpretable classifiers. Furthermore, generalized probabilistic datasets correspond to probabilistic databases, and so advances in this field will also enhance research in this direction.
\revision{In the Appendix~\ref{appendix:discussion_interpretation}, we further discuss how $\newmetric$
can be interpreted and how it relates to other information theory metrics.}

\revision{Interestingly, our approach can be seen as upper-bounding the capacity of any adversary (with the exact same information) for reconstructing the training dataset. Indeed, each (deterministic) reconstruction encoded via our (generalized) probabilistic dataset is compatible with the trained model’s structure, hence without further information no adversary should be able to distinguish between them.
The proposed approach could also be leveraged to mount further inference attacks targeting the model's training data~\cite{DBLP:journals/corr/abs-2007-07646}. For instance, one could search within the encoded set of possible reconstructions whether a particular profile appears. If it never appears, then the result of the associated membership inference attack is certifiably negative. If it does, the proportion of possible reconstructions in which it exists can be used as an estimate of the probability of inclusion in the training data. Property inference attacks could also be conducted by averaging statistics over the set of possible reconstructions.}

A promising extension of our study consists in leveraging the knowledge of the learning algorithm's internals to lower the reconstructed generalized probabilistic dataset entropy. 
For instance, if a greedy algorithm uses the Gini impurity as a splitting criterion, we know that at a given node no feature other than the chosen one can yield a better Gini impurity value in the training set.
Additionally, optimality itself gives information: some combinations of the attributes not used within an optimal decision tree can be discarded if they could allow the building of a better decision tree. 
\revision{General demographic information could also help identify more probable reconstructions. Technically speaking, external knowledge could either discard some of the possible reconstructions or modify the probability distribution over them. Importantly, we do not consider it in this paper in order to remain algorithm-agnostic and focus on the interpretable models themselves.}

We empirically observed that the entropy reduction brought by the knowledge of some interpretable model is not uniform across all examples. 
Investigating whether it disproportionately affects some subgroup of the population is an interesting direction.
Another promising future work consists in combining the knowledge of different generalized probabilistic datasets, as was proposed in~\cite{DBLP:conf/dbsec/GambsGH12}. 
This would require aligning them, as well as merging several probability distributions, while in the original setup it simply consisted in union of sets.
\revision{Finally, learning interpretable models optimizing our information leak measure could also constitute an interesting defense mechanism, in particular to avoid worst-case leakages as identified in our experiments.}
Investigating the effect of privacy-preserving methods such as the widely used \emph{Differential Privacy}~\cite{dwork2014algorithmic} on the quality of the probabilistic reconstruction 
is also an insightful research avenue. \revision{In the Appendix~\ref{appendix:reconstruction_from_dp}, we overview challenges related to attacking differentially private models and discuss possible techniques to address them.}

\bibliographystyle{IEEEtran}
\bibliography{IEEEabrv,references}

\appendices

\section{Additional Results}
\label{appendix:additional_results_p_datasets}

We observed in Section~\ref{sec:results} that optimal models (either decision trees or rule lists) usually contain more information than greedily-built ones of the same size.
However, when related to the models' utility (accuracy on the training data), this trend is reversed, and optimal models leak less information regarding their training data than greedily-built ones for the same performances level.
This was explained by the fact that greedy learning algorithms iteratively make local choices that are sub-optimal, overall adding unnecessary information to the resulting model (\emph{e.g.,} performing more splits than necessary within decision trees).

\begin{figure*}[htb!]
     \begin{center}
    
    \begin{subfigure}{\textwidth}
        \centering
        
        \includegraphics[width=\figsize\textwidth]{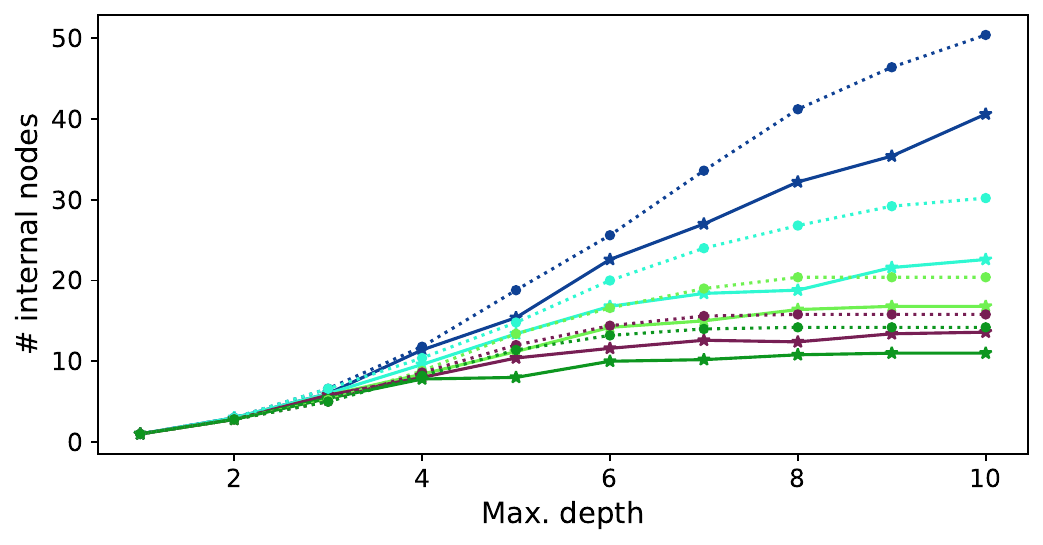}
        \includegraphics[width=\figsize\textwidth]{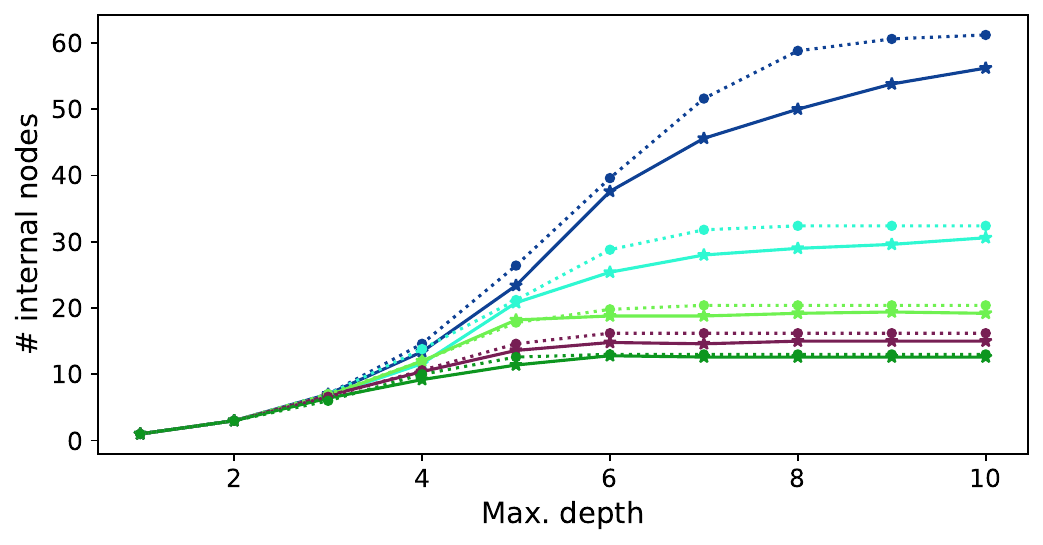}
   
        \caption{Experiments relating the actual models' sizes to the size constraints enforced during learning. 
        We report tree size (number of splits/internal nodes) as a function of the maximum depth constraint.~\label{fig:results_trees_n_elementary_tokens_f_max_depth}}
        
    \end{subfigure}
    
    \par\bigskip 
       
        \begin{subfigure}{\textwidth}
        \centering
        
        \includegraphics[width=\figsize\textwidth]{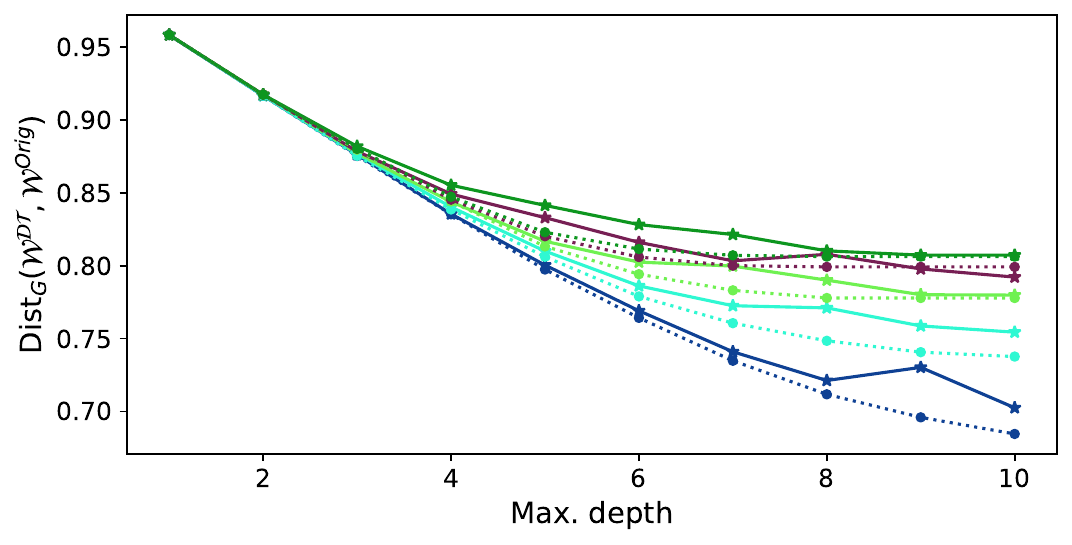}
        \includegraphics[width=\figsize\textwidth]{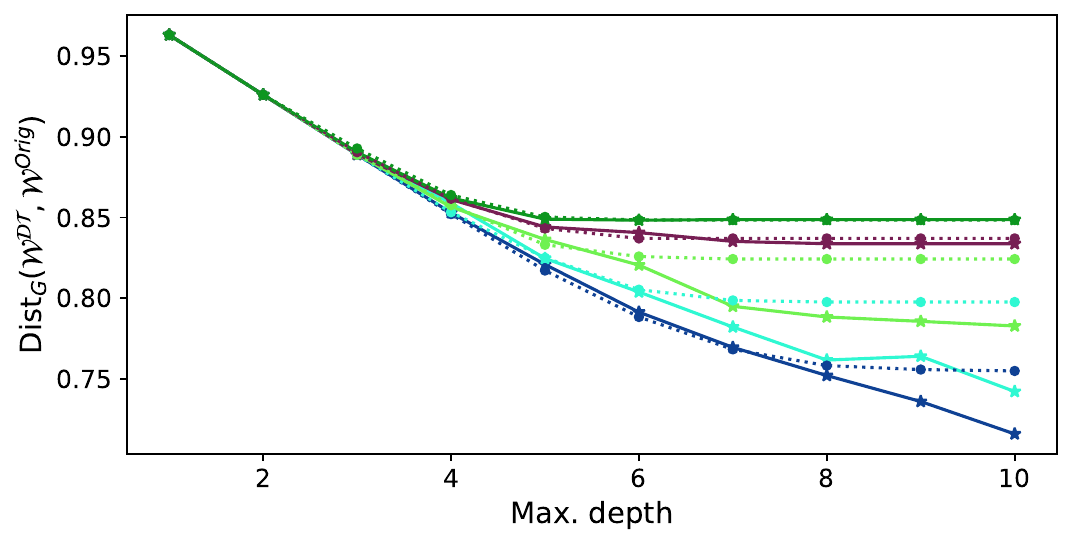}
   
        \caption{Experiments relating the entropy reduction ratio to the size constraints enforced during learning.
        We report the entropy reduction as a function of the maximum depth constraint.~\label{fig:results_trees_entropy_reduction_f_max_depth}}
        
    \end{subfigure}
    
     \vskip 10 pt

    \includegraphics[width=\legsize\textwidth]
   {dts_results_legend.pdf}
   
     \vskip 10 pt
    \caption{Results of our experiments comparing optimal and greedily-built decision trees (learnt respectively with \exactdt{} and \heuristicdt{}), for different (relative) minimum leaf support values.
    Left: Adult Income dataset, Right: COMPAS dataset.}  
    \label{fig:results_models_size_trees}
    \end{center}
\end{figure*}

\begin{figure*}[htb!]
     \begin{center}

     \begin{subfigure}{\textwidth}
        \centering
        
        \includegraphics[width=\figsize\textwidth]{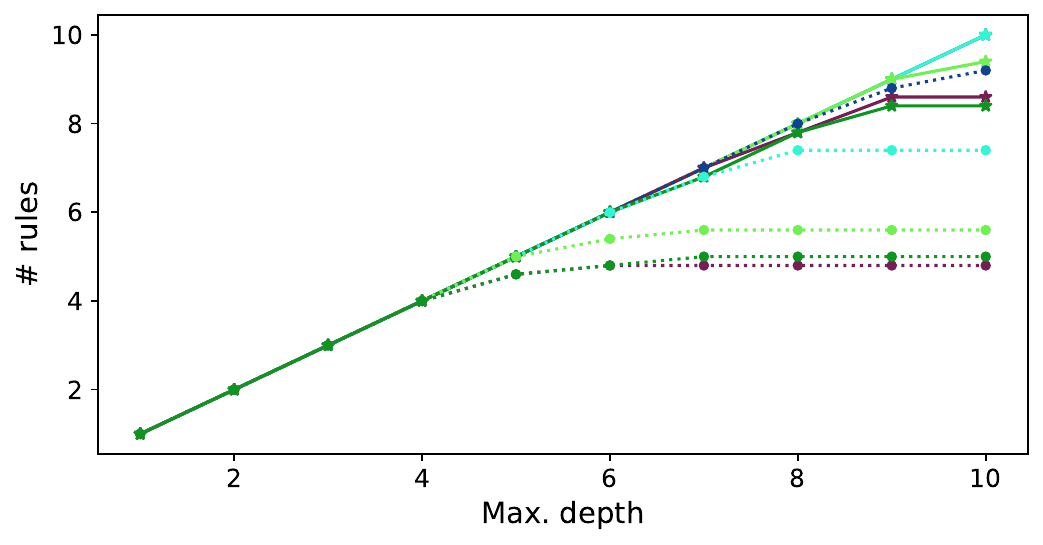}
        \includegraphics[width=\figsize\textwidth]{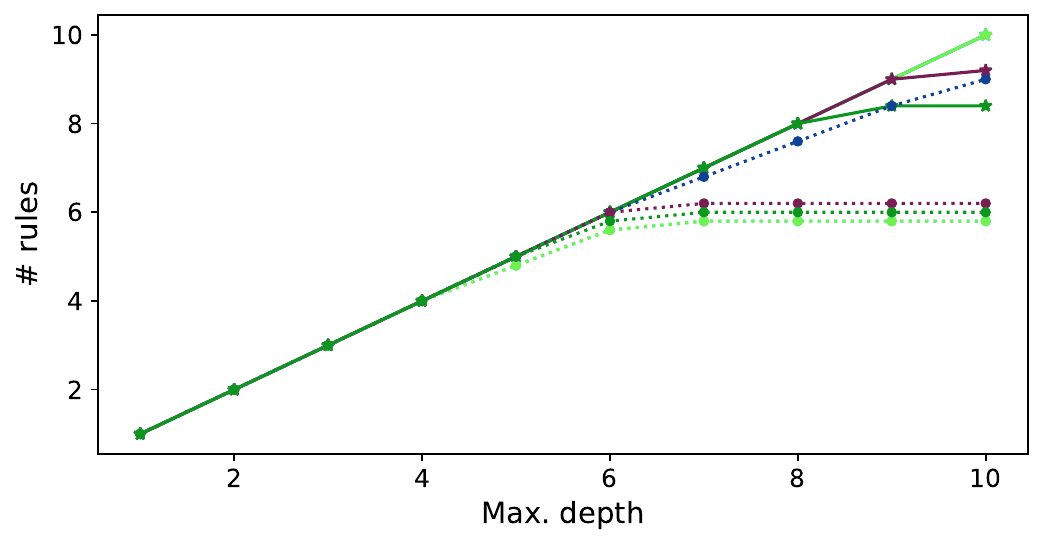}

        \caption{Experiments relating the actual models' sizes to the size constraints enforced during learning. 
        We report rule list size (number of rules) as a function of the maximum depth constraint.~\label{fig:results_rule_lists_n_elementary_tokens_f_max_depth}}
        
    \end{subfigure}
    
    \par\bigskip 
        \begin{subfigure}{\textwidth}
        \centering
        
        \includegraphics[width=\figsize\textwidth]{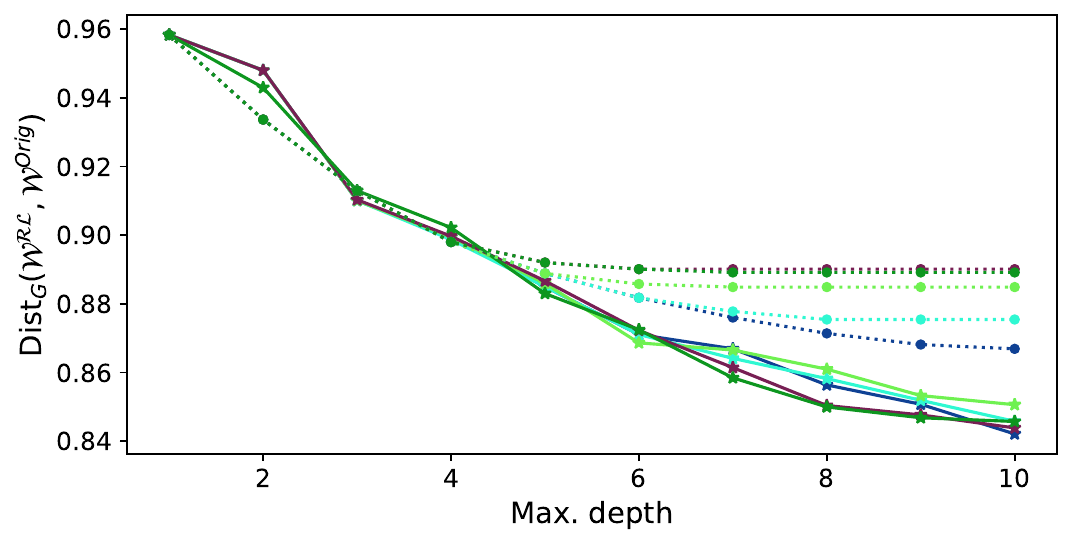}
        \includegraphics[width=\figsize\textwidth]{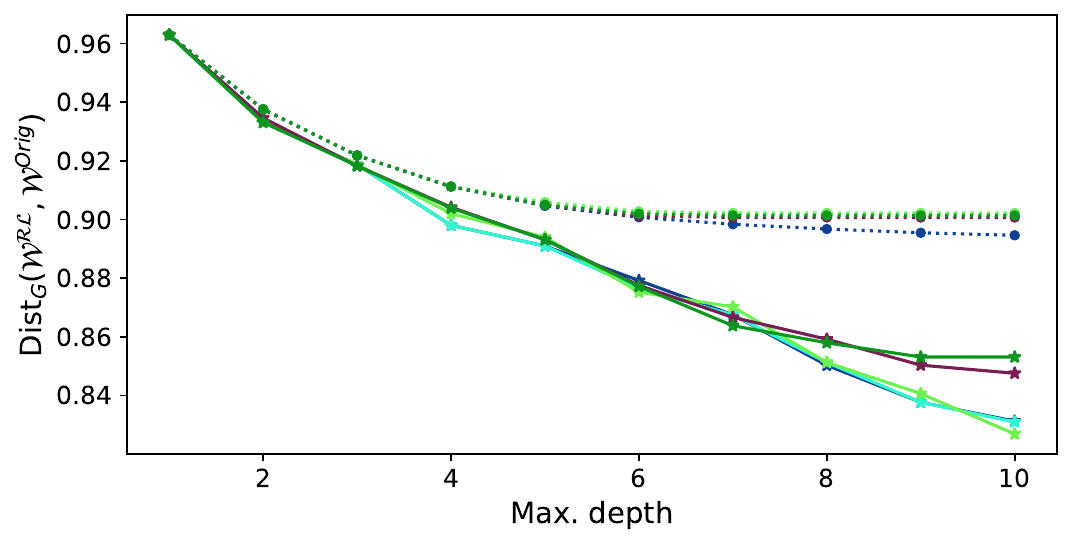}
   
        \caption{Experiments relating the entropy reduction ratio to the size constraints enforced during learning. 
        We report the entropy reduction as a function of the maximum depth constraint.~\label{fig:results_rule_lists_entropy_reduction_f_max_depth}}
        
    \end{subfigure}
    
    \vskip 10 pt

    \includegraphics[width=\legsize\textwidth]
   {rls_results_legend.pdf}
   
    \caption{Results of our experiments comparing optimal and greedily-built rule lists (learnt respectively with \corels{} and \greedyRL{}), for different (relative) minimum rule support values.    
    Left: Adult Income dataset, Right: COMPAS dataset.}  
    \label{fig:results_models_size_rule_lists}
    \end{center}
\end{figure*}

In this appendix section, we relate the amount of information an interpretable model carries to the size constraints that were enforced to build it.
More precisely, we report in Figures~\ref{fig:results_trees_n_elementary_tokens_f_max_depth} and~\ref{fig:results_rule_lists_n_elementary_tokens_f_max_depth} the resulting model size as a function of the maximum depth constraint, for the different minimum support constraints.
Again, the model size is quantified as the number of internal nodes for a decision tree, or as the number of rules for width-1 rule lists.
We also report in Figures~\ref{fig:results_trees_entropy_reduction_f_max_depth} and~\ref{fig:results_rule_lists_entropy_reduction_f_max_depth} the overall entropy reduction ratio, as a function of the maximum depth constraint.

One can observe in Figure~\ref{fig:results_trees_n_elementary_tokens_f_max_depth} that, as expected, the number of internal nodes within the built decision trees grows with the maximum depth value.
Enforcing large values of the (relative) minimum leaf support quickly prevents the trees from expanding, as no split can be performed while satisfying the minimum support constraint.
Hence, as expected, lowering the minimum support value leads to the computation of larger decision trees.
Comparing greedily-built and optimal decision trees, one can note that the models learnt by \heuristicdt{} contain more nodes than the optimal ones built using \exactdt{}, for the same provided parameters (\emph{i.e.,} minimum leaf support and maximum depth values).
This can be explained by the fact that \heuristicdt{} often adds non-necessary splits as it iteratively performs local, sub-optimal choices. 
Meanwhile, many branches do not reach the enforced maximum depth within the optimal decision trees thanks to the performed global optimization which considers a split only if it is necessary. 
As a consequence, we observe in Figure~\ref{fig:results_trees_entropy_reduction_f_max_depth} (left) that, for fixed parameters, the decision trees produced by \heuristicdt{} on the Adult Income dataset contain more information than those learnt by \exactdt{}.
For the COMPAS dataset (Figure~\ref{fig:results_trees_entropy_reduction_f_max_depth} (right)), we observe the opposite trend. 
This can be explained by two observations.
First, the size difference between optimal and greedily-built decision trees is smaller on the COMPAS dataset (Figure~\ref{fig:results_trees_n_elementary_tokens_f_max_depth} (right)) than on the Adult dataset (Figure~\ref{fig:results_trees_n_elementary_tokens_f_max_depth} (left)).
Then, in average, we saw within Section~\ref{sec:results} (Figure~\ref{fig:results_trees_entropy_reduction_f_n_elementary_tokens}) that, for equivalent sizes, the optimal decision trees carry more information than the greedily-built ones.

Figure~\ref{fig:results_rule_lists_n_elementary_tokens_f_max_depth} shows that, as expected, the number of rules within the built rule lists grows with the enforced maximum depth value.
As for the decision trees, largest values of the enforced minimum rule support prevent expansion of the rule lists, when no rule satisfying the minimum support constraint can be found.
This is particularly true for the greedy learning algorithm. Indeed, at each iteration, the algorithm selects a rule maximizing a given criterion (\emph{i.e.,} minimizing Gini Impurity). 
Then, the examples not captured by the rules fall into the rest of the rule list, and are used for the next iterations.
If the algorithm selects rules with large supports during the first iterations, there may be too few remaining examples to be able to add new rules.
This drawback is not observed with \corels{}, as it performs global optimization.
As a direct consequence, one can see in Figure~\ref{fig:results_rule_lists_entropy_reduction_f_max_depth} that, for fixed parameters (\emph{i.e.,} minimum rule support and maximum depth values), the rule lists built using \corels{} contain more information than those produced by \greedyRL{}. This trend is related to the observed size difference, but is also exacerbated by the fact that, as observed in Section~\ref{sec:results} (Figure~\ref{fig:results_rule_lists_entropy_reduction_f_n_elementary_tokens}), optimal rule lists usually encode more information that greedily-built ones of equivalent size.

\begin{figure*}[htb!]
     \begin{center}
    
     \begin{subfigure}{\textwidth}
        \centering
        
        \includegraphics[width=\figsize\textwidth]{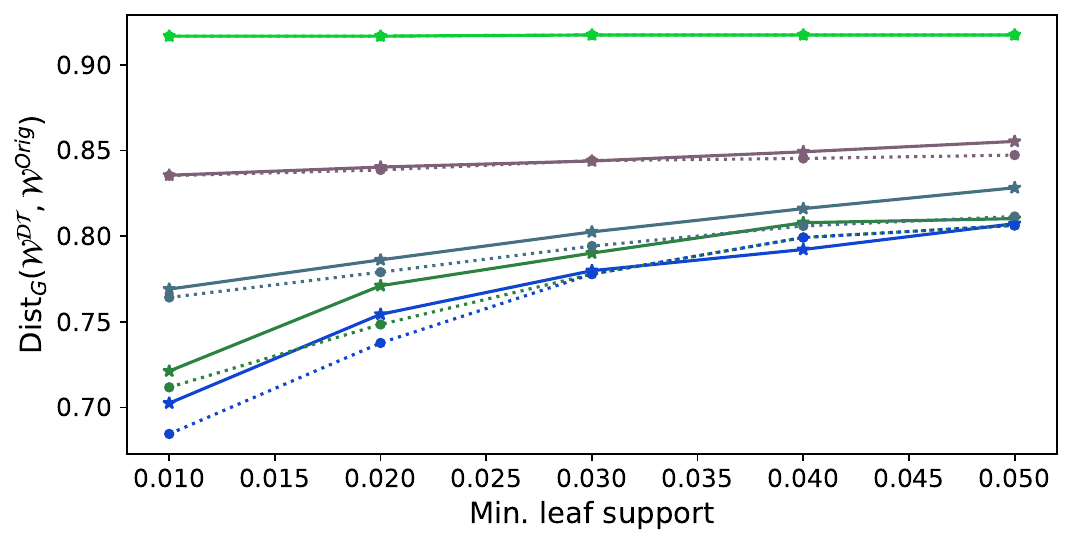}
        \includegraphics[width=\figsize\textwidth]{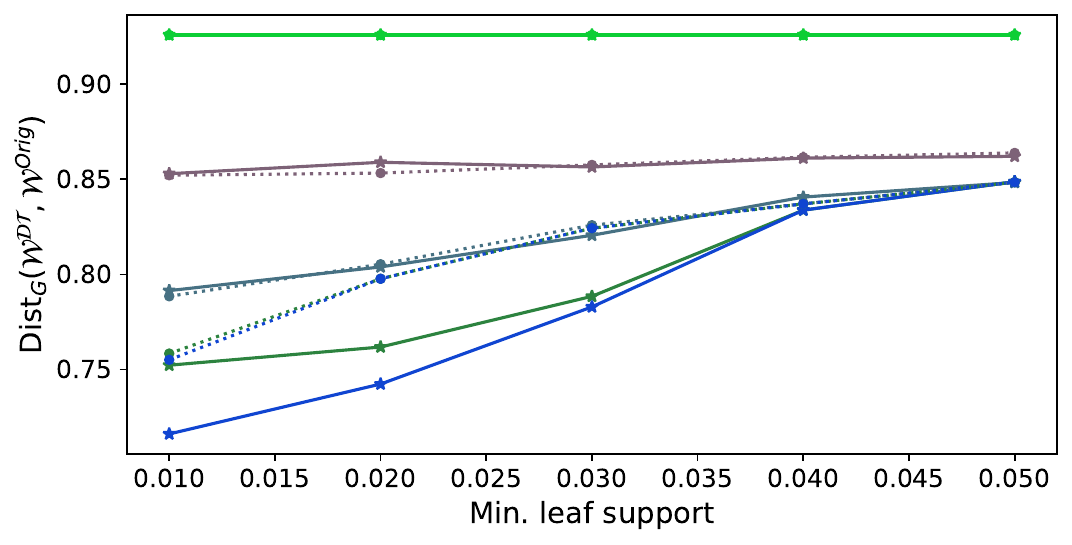}

        \includegraphics[width=\legsize\textwidth]
   {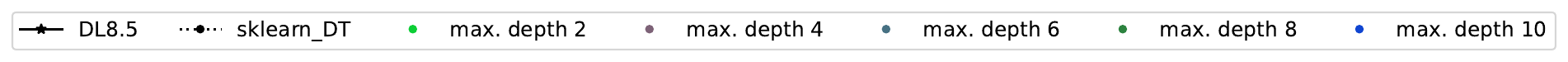}
   
        \caption{\revisionmultiline{
        Optimal and greedily-built decision trees, learned respectively with the \exactdt{} and \heuristicdt{} algorithms.}~\label{fig:results_trees_entropy_reduction_f_min_support}}
        
    \end{subfigure}

    \par\bigskip 
    
     \begin{subfigure}{\textwidth}
        \centering
        
        \includegraphics[width=\figsize\textwidth]{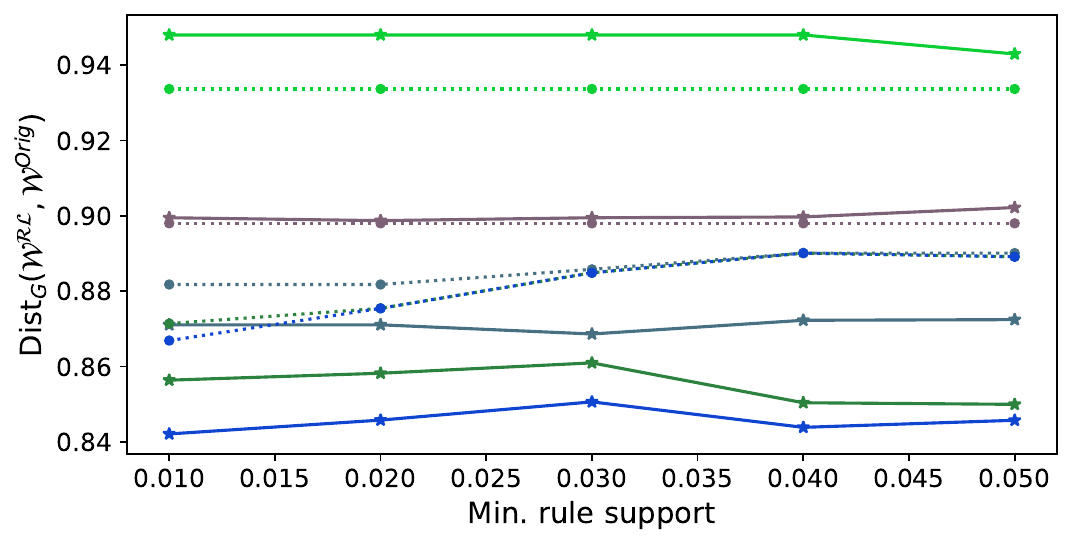}
        \includegraphics[width=\figsize\textwidth]{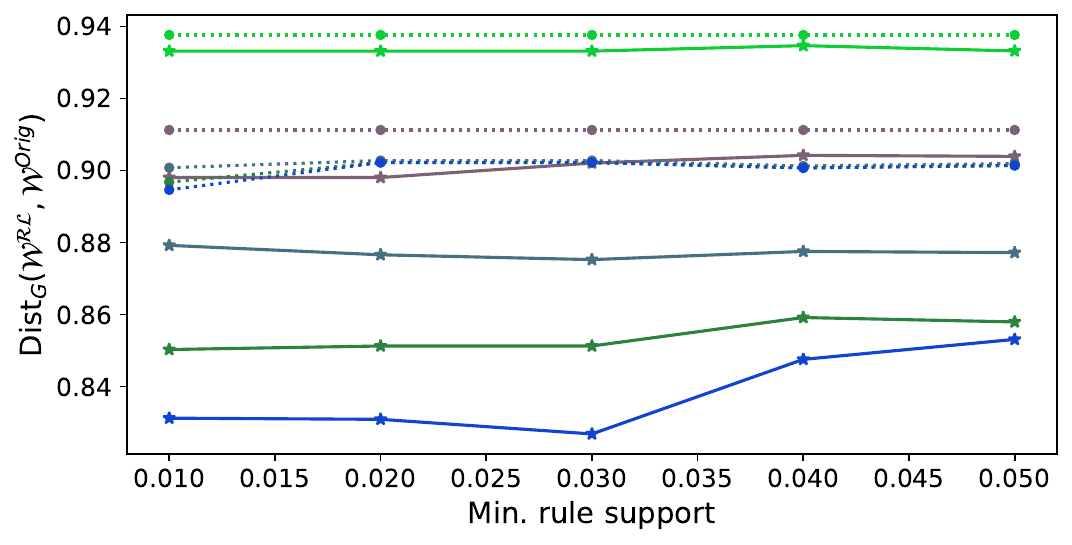}

        \includegraphics[width=\legsize\textwidth]
   {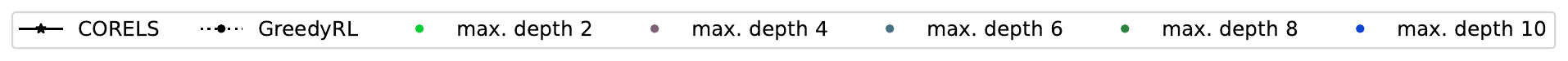}
   
        \caption{\revisionmultiline{
        Optimal and greedily-built rule lists, learned respectively with the \corels{} and \greedyRL{} algorithms.}~\label{fig:results_rules_entropy_reduction_f_min_support}}
        
    \end{subfigure}
   
    \caption{\revisionmultiline{
    Results of our experiments relating the entropy reduction ratio to the (relative) minimum support constraints enforced during learning.
    We report the entropy reduction as a function of the (relative) minimum (leaf or rule) support constraint, for different maximum depth constraint values.
    Left: Adult Income dataset, Right: COMPAS dataset.}}  
    \label{fig:results_both_entropy_reduction_f_min_support}
    \end{center}
\end{figure*}

\revisionmultiline{
Finally, we report in Figure~\ref{fig:results_both_entropy_reduction_f_min_support} the entropy reduction ratio as a function of the minimum (leaf or rule) support constraint, for optimal and greedily-built decision trees (Figure~\ref{fig:results_trees_entropy_reduction_f_min_support}) or rule lists (Figure~\ref{fig:results_rules_entropy_reduction_f_min_support}).
For decision tree models, one can observe that the minimum leaf support value only influences learning for large enough maximum depth constraints. In such case, smaller minimum support values allow the building of longer branches, which results in decision trees containing more information.
The trend is not that clear for rule lists models, which is explained by two observations. 
First, contrary to decision trees, rule lists exactly matching the maximum depth constraint can almost always be built using \corels{}, which performs global optimization, as can be observed in Figure~\ref{fig:results_rule_lists_n_elementary_tokens_f_max_depth}. In such a case, the minimum support value does not actually influence the size of the learnt model.
Second, while the leaves of a decision tree have disjoint supports, this is not the case for the successive rules within a rule list, and enforcing greater support for the first rules impacts the possible support values for the following ones.
Overall, the relationship between the minimum rule support constraint and the entropy reduction ratio is complex, and no clear trend can be observed. On the opposite, the minimum leaf support constraint used when building a decision tree strongly influences the tree size and the amount of information it contains.
}


\revisionmultiline{
\section{Discussion on the Proposed Attack Success Metric}
\label{appendix:discussion_interpretation}

The objective of the considered adversary is to reconstruct the entire training dataset $\generalizedpdataset^{\origd}$, \emph{i.e.,} all attributes of all examples: $w^{\origd}_{\exampleindex \in [1..\nsamples], \attributeindex \in [1..\nfeatures]}$.
Thus, from the perspective of an uninformed adversary (who only knows the training set size $\nsamples$ and the domains $\singledomain_\attributeindex$ of the different attributes $\attributeindex \in [1..\nfeatures]$) all the possible combinations of all attributes' values (within their domains) are possible for every example.

Any knowledge reducing the scope of possible reconstructions hence eases the adversary's task and raises the probability of correct reconstruction if the adversary chooses among the (reduced) set of possible reconstructions. This reduction is precisely what our proposed metric, $\newmetric(\generalizedpdataset^{\interpretablemodel},\generalizedpdataset^{\origd})$, quantifies, through the ratio of the (joint) entropy of the (generalized) probabilistic dataset over the (joint) entropy of the uninformed reconstruction. Intuitively, it measures the uncertainty remaining in the built (generalized) probabilistic reconstruction $\generalizedpdataset^{\interpretablemodel}$, through the joint entropy of all the random variables $\onecell[\exampleindex \in [1..\nsamples], \attributeindex \in [1..\nfeatures],\generalizedpdataset^{\interpretablemodel}]$ modeling the dataset. As all reconstructions are compatible with the provided model’s structure, and the true (unknown) training dataset is provably part of the encoded set of possible reconstructions, this uncertainty measure is a good indicator on how much we can reduce the space of possible solutions around the correct one. In other words, $\newmetric$ can be seen as a measure of reconstructibility for the whole dataset.

One could note that the mutual information between the input database and the (generalized) probabilistic reconstruction may be an interesting indicator. However, its computation would reduce to the probability of the true input database in the built (generalized) probabilistic reconstruction $\mathbb{P}\left(\onecell[\exampleindex \in [1..\nsamples],\attributeindex \in [1..\nfeatures],\generalizedpdataset] = \world^{\origd}_{\exampleindex \in [1..\nsamples],\attributeindex \in [1..\nfeatures]}\right)$. While this indicator would also be insightful, it would ignore the other reconstructions that are compatible with the model’s structure. In practice, interpretable models (in particular, decision trees and rule lists) allow the building of a set of possible reconstructions, but without further information, one cannot distinguish between them. In a sense, they are all equally probable and the probability of the correct reconstruction is exactly $\frac{1}{\lvert \possibleworlds(\generalizedpdataset{}^{\interpretablemodel}) \rvert}$, which goes back to our proposed computation (Section~\ref{sec:reconstruction_in_practice}).
Formulating other reconstructibility criteria leveraging tools from the information theory is an interesting research avenue, building upon our thorough formalization of the notion of (generalized) probabilistic datasets.
}
\revisionmultiline{
\section{(Generalized) Probabilistic Reconstruction from Differentially Private Models: Challenges}
\label{appendix:reconstruction_from_dp}

Assessing the effect of Differential Privacy (DP)~\cite{dwork2014algorithmic} on the proposed probabilistic reconstruction would be insightful. 
This would require building interpretable models using differentially private learning algorithms.
In particular, varying the privacy budget and observing the (potential) impact on the resulting probabilistic reconstruction success would be interesting to assess whether (and for which parameters) DP provides a valuable protection against such dataset-level attacks.

For instance, one could consider differentially private decision trees~\cite{10.1145/1835804.1835868,DBLP:journals/csur/FletcherI19}. Note that the rationale of the following discussion also holds for rule lists or other types of interpretable models.
Recall that the aim of our approach is to construct an object (a (generalized) probabilistic dataset) encoding the whole set of dataset reconstructions that are compatible with the structure of a given (trained) model.
One key challenge is that DP decision trees can only provide noisy counts within the tree leaves. Indeed, such counts are required to compute confidence scores, but the exact per-class cardinalities (as displayed within the decision tree represented in Figure~\ref{fig:example_toy_dt}) would violate DP. These noisy counts, if used directly in the proposed iterative reconstruction process (which follows the path through each branch to perform the attributes’ domains reductions as specified by the internal nodes’ splits), would then induce errors in the performed reconstruction. In the worst case, the true reconstruction could not be part of the possible ones (represented using the constructed (generalized) probabilistic dataset) due to these errors. In such a case, one can still use the (generalized) probabilistic dataset variables’ joint entropy to estimate the uncertainty within the encoded set of possible reconstructions, but measuring the distance between the true dataset and the set of plausible reconstructions given the DP model would also be an interesting indicator.

Considering other types of models (such as rule sets), DP could even lead to infeasible reconstruction problems (\emph{i.e.,} empty set of possible reconstructions) due to inconsistent (noisy) counts within the different rules. 
A promising mitigation technique would be to account for the added noise, for instance by modeling the noise values using linear programming as was done in previous works~\cite{DBLP:journals/jpc/CohenN20}. However, this requires the use of more elaborated techniques and is out of the scope of this paper. 
}

\end{document}